\documentclass[sigconf]{acmart}

\usepackage{bbm}
\usepackage{subcaption}
\usepackage{amsmath}
\usepackage{algorithmic}
\usepackage{multirow}
\usepackage{pifont}
\usepackage{bbding}
\usepackage{enumitem}
\usepackage{amsthm}
\usepackage{titletoc}
\usepackage{diagbox}
\usepackage{makecell}
\usepackage{tabularx}
\usepackage{wrapfig}

\AtBeginDocument{%
  }

\setcopyright{acmlicensed}
\copyrightyear{2018}
\acmYear{2018}
\acmDOI{XXXXXXX.XXXXXXX}
\acmConference[Conference acronym 'XX]{Make sure to enter the correct
  conference title from your rights confirmation email}{June 03--05,
  2018}{Woodstock, NY}
\acmISBN{978-1-4503-XXXX-X/2018/06}

\begin{document}
\graphicspath{{figures/}}

\title{Personalized Federated Fine-Tuning for LLMs via Data-Driven Heterogeneous Model Architectures}

\author{Yicheng Zhang}
\affiliation{%
  \institution{Zhejiang University}
  \city{Hangzhou}
  \state{Zhejiang}
  \country{China}}
\email{zyccs@zju.edu.cn}

\author{Zhen Qin}
\authornote{Zhen Qin and Shuiguang Deng are co-corresponding authors.}
\affiliation{%
  \institution{Zhejiang University}
  \city{Hangzhou}
  \state{Zhejiang}
  \country{China}}
\email{zhenqin@zju.edu.cn}

\author{Zhaomin Wu}
\affiliation{%
  \institution{National University of Singapore}
  \country{Singapore}}
\email{zhaomin@nus.edu.sg}

\author{Jian Hou}
\affiliation{%
  \institution{Zhejiang Sci-Tech University}
  \city{Hangzhou}
  \state{Zhejiang}
  \country{China}}
\email{changeleap@163.com}

\author{Shuiguang Deng}
\authornotemark[1]
\affiliation{%
  \institution{Zhejiang University}
  \city{Hangzhou}
  \state{Zhejiang}
  \country{China}}
\email{dengsg@zju.edu.cn}

\renewcommand{\shortauthors}{Trovato et al.}

\begin{abstract}
Large language models (LLMs) are increasingly powering web-based applications, whose effectiveness relies on fine-tuning with large-scale instruction data.
However, such data often contains valuable or sensitive information that limits its public sharing among business organizations.
Federated learning (FL) enables collaborative fine-tuning of LLMs without accessing raw data.
Existing approaches to federated LLM fine-tuning usually adopt a uniform model architecture, making it challenging to fit highly heterogeneous client-side data in varying domains and tasks, e.g., hospitals and financial institutions conducting federated fine-tuning may require different LLM architectures due to the distinct nature of their domains and tasks.
To address this, we propose FedAMoLE, a lightweight personalized FL framework that enables data-driven heterogeneous model architectures. It features a heterogeneous mixture of low-rank adaptation (LoRA) experts module to aggregate architecturally heterogeneous models and a reverse selection-based expert assignment strategy to tailor model architectures for each client based on data distributions. Experiments across seven scenarios demonstrate that FedAMoLE improves client-side performance by an average of 5.97\% over existing approaches while maintaining practical memory, communication, and computation overhead.
\end{abstract}

\ccsdesc[500]{Computing methodologies~Distributed artificial intelligence}
\ccsdesc[500]{Computing methodologies~Learning paradigms}

\keywords{Personalized Federated Learning, Large Language Models, Heterogeneous Model Architectures, Mixture of LoRA Experts.}

\received{20 February 2007}
\received[revised]{12 March 2009}
\received[accepted]{5 June 2009}

\maketitle

\section{Introduction}
Large language models (LLMs) are profoundly reshaping the contemporary web experience, serving as the core engine of many web-based applications, such as recommender systems \cite{gao2025llm4rerank}, content analysis \cite{qiao2025thematic}, and web agents \cite{ning2025survey}.
The performance of these applications directly depends on the capabilities of the underlying LLMs.
Instruction tuning, as a key step in shaping the exceptional capabilities of LLMs, requires large amounts of instruction-based text data, as suggested by the scaling law~\cite{kaplan2020scaling}. 
As publicly available data gradually depletes \cite{villalobos2024position}, sharing instructional data between business organizations to jointly enhance the capabilities of LLMs will effectively build a competitive advantage, since business organizations typically accumulate a large amount of instructional data while operating businesses powered by LLMs.
However, the privacy information and commercial value contained in instructional data somewhat restrict the direct sharing of such data.

Federated learning (FL)~\cite{mcmahan2017fl} enables multiple business organizations to collaboratively train models without sharing raw data. 
Recently, it has increasingly been used to mine the value of client-side data in a privacy-preserving manner to enhance LLM capabilities\cite{kuangFederatedScopeLLMComprehensivePackage2023,qinFederatedFullParameterTuning2023b,qin2025federated}.
Given the large scale of LLMs, recent efforts on federated fine-tuning mainly focus on improving system efficiency~\cite{zhangBuildingFederatedgptFederated2023,cheFederatedLearningLarge2023,qinFederatedFullParameterTuning2023b}, typically following standard FL settings that produce a single globally consistent model.
However, in cross-silo scenarios where different business organizations hold data from diverse domains or following various distributions, client-side data is often statistically heterogeneous (a.k.a. non-independent and identically distributed, non-IID). 
This makes it difficult for a single global model to generalize across all client-side data, resulting in suboptimal accuracy~\cite{tanPersonalizedFederatedLearning2023}.

\begin{figure}[t]
    \centering
    \includegraphics[width=0.94\linewidth]{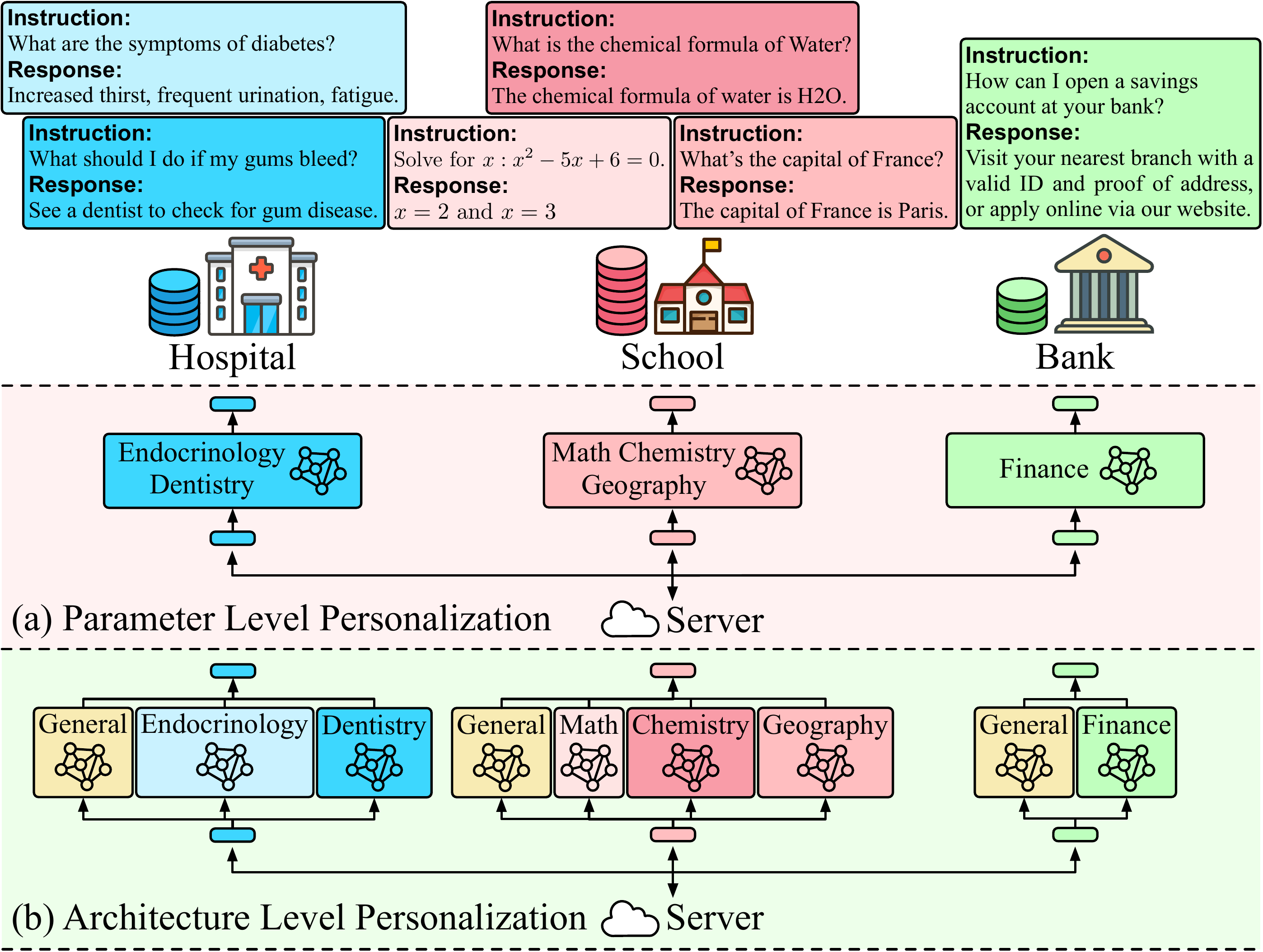}
    \caption{Scenario examples of federated LLM fine-tuning among three institutes, where each holds data that differs in domain and tasks.
    To better adapt to local data, personalized models with heterogeneous architectures tailored for local data distributions are usually preferred (b).}
    \label{fig:scene}
\end{figure}
\begin{figure}[t]
    \centering
    \begin{subfigure}[t]{0.49\linewidth}
        \centering
        \includegraphics[width=\linewidth]{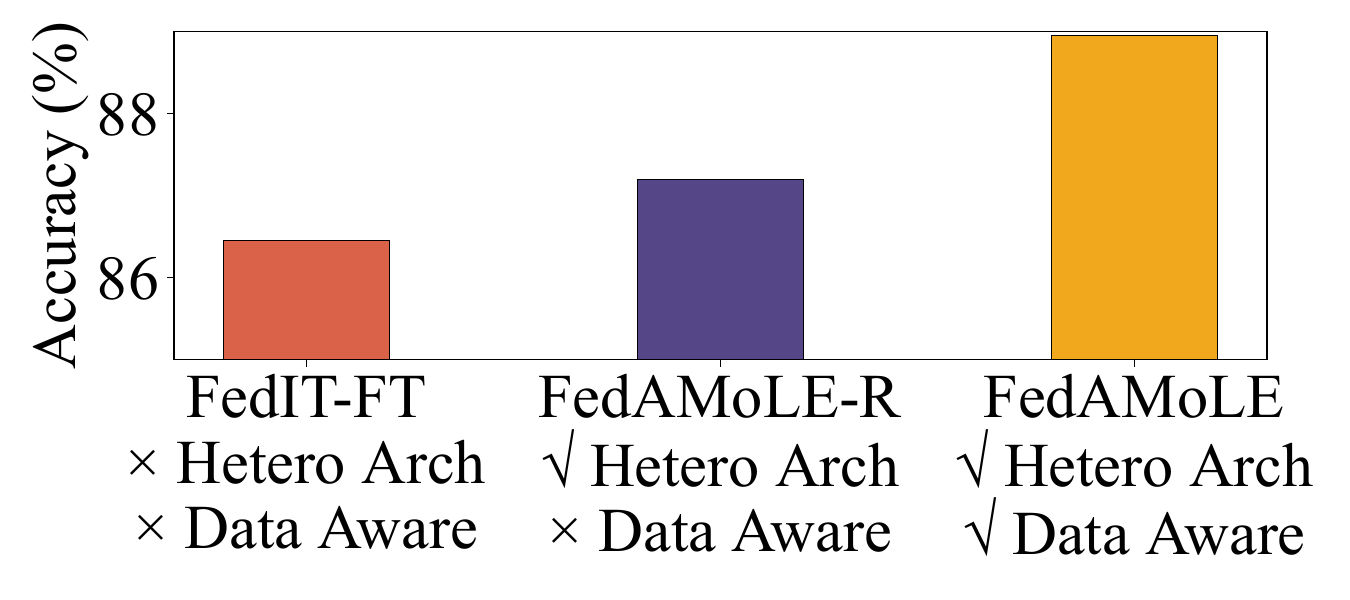}
        \caption{SNLI}
        \label{ablation_bar_snli}
    \end{subfigure}
    \begin{subfigure}[t]{0.49\linewidth}
        \centering
        \includegraphics[width=\linewidth]{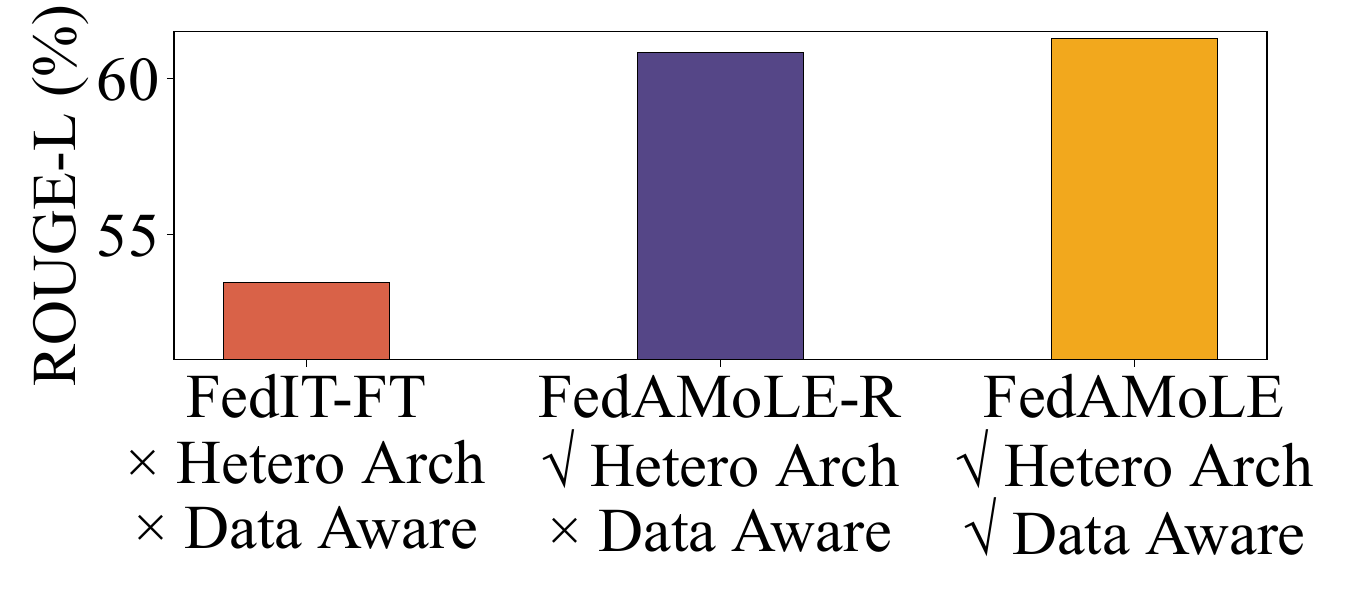}
        \caption{Natural Instructions}
        \label{ablation_bar_ni}
    \end{subfigure}
    \caption{Effectiveness of heterogeneous and data-driven architectures. FedIT-FT and FedAMoLE-R use homogeneous and heterogeneous architectures (Hetero Arch), respectively. FedAMoLE further adopts data-driven architecture optimization. (Please refer to \S\ref{sec:ablation} for more details)}
    \label{fig:ablation_bar}
\end{figure}
Prior FL studies indicate that enabling multiple personalized models can improve performance on client-side data~\cite{linEnsembleDistillationRobust2020,qin2023fedapen}.
However, due to their dependence on external proxy data used for knowledge distillation, or otherwise high resource costs, some existing personalized FL (PFL) methods are impractical for fine-tuning LLMs (as discussed in \S\ref{sec:pfl_related_work}).
Recent studies have explored PFL for LLMs through distillation-free personalization~\cite{jiangPersonalizedWirelessFederated2024, yangDualpersonalizingAdapterFederated2024,qiFDLoRAPersonalizedFederated2024}. 
Despite their promising results, they still face two key limitations:
\begin{itemize}[leftmargin=*]
    \item (L1) \textbf{Limited Support for Architecture-Level Personalization:} Most existing PFL methods for LLM fine-tuning focus primarily on parameter-level personalization within a fixed global architecture, as illustrated in Figures~\ref{fig:scene}(a)~\cite{qiFDLoRAPersonalizedFederated2024, almansooriCollaborativeEfficientPersonalization2024}.
    This constraint reduces the flexibility of adaptation across highly non-IID client data.
    As shown in Figure~\ref{fig:ablation_bar}, the method that enables architectural heterogeneity through personalized submodule (expert) compositions, i.e., FedAMoLE-R, consistently outperforms another PFL approach that relies on a fixed shared architecture (FedIT-FT).
    Thus, it is important to explore architecture-level personalization in federated LLM fine-tuning.
    
    \item (L2) \textbf{Lack of Structural Adaptation to Local Data:} 
    Existing FL methods supporting heterogeneous LLMs are mainly designed with manually predefined model architectures~\cite{jiangPersonalizedWirelessFederated2024,yangDualpersonalizingAdapterFederated2024,zhangPersonalizedFederatedInstruction2024}, which \emph{lack joint awareness of intra- and inter-client data distributions}.
    As shown in Figure~\ref{fig:scene}(b), FedAMoLE (ours) optimizes personalized model architectures based on data distributions to align experts with client-side data domains, thereby improving fine-tuning accuracy (Figure \ref{fig:ablation_bar}).
    Thus, exploring data-aware model architectures is crucial for LLM fine-tuning in FL.
\end{itemize}

This work aims to enable the tailoring of personalized model architectures informed by client-side data distributions.
To this end, we propose FedAMoLE, a PFL framework for fine-tuning personalized LLMs that supports adaptation to local data distributions through data-driven heterogeneous model architectures.
It consists of two key components:
1) A \textbf{h}eterogeneous \textbf{m}ixture \textbf{o}f \textbf{l}ow-rank adaptation \textbf{e}xpert (\textbf{HMoLE}) module, which employs a novel shape-invariant router to \textbf{enable aggregation of architecturally heterogeneous models} (L1).
2) A \textbf{r}everse \textbf{s}election-based \textbf{e}xpert \textbf{a}ssignment (\textbf{RSEA}) strategy for \textbf{expert assignment}, where experts select domain-relevant clients from a global perspective, enabling data-driven architecture optimization based on client-side data distributions without direct access to local data (L2).
The main contributions of this work are as follows:
\begin{itemize}[leftmargin=*]
    \item We propose FedAMoLE, a novel PFL framework for fine-tuning LLMs that enables data-driven heterogeneous model architectures. It features a module that supports heterogeneous model aggregation and a strategy for client model customization, making it uniquely capable of adapting to diverse client-side data.
    \item To aggregate architecturally heterogeneous models, we design HMoLE module for FedAMoLE, which employs a shape-invariant router to coordinate a heterogeneous mixture of LoRA experts.
    \item To tailor model structures for client-side data, we propose RSEA, a data-driven architecture optimization strategy in FedAMoLE. It adopts a novel expert-to-client selection mechanism that jointly considers intra- and inter-client data characteristics.
    \item We conduct extensive experiments across seven non-IID FL scenarios, showing that FedAMoLE improves task accuracy by an average of 5.97\% over the strongest baseline. Our codes are publicly available at 
    \url{https://github.com/zyc140345/FedAMoLE}.
\end{itemize}
\section{Related Work}
This section briefly overviews existing studies highly related to this work. 
For detailed discussions, please refer to Appendix \ref{appendix:related-work}.

\emph{Personalized Federated LLM Fine-Tuning}
Currently, most FL methods for LLM fine-tuning adopt Parameter-Efficient Fine-Tuning (PEFT) techniques such as Low-Rank Adaptation (LoRA)~\cite{zhangBuildingFederatedgptFederated2023,wuFedBiOTLLMLocal2024,qin2025federated} and prompt tuning~\cite{cheFederatedLearningLarge2023} to reduce resource costs.
These methods typically yield a single global model that fails to adapt to heterogeneous client-side data. Recent studies explore personalization via dual adapters~\cite{qiFDLoRAPersonalizedFederated2024}, LoRA masking~\cite{zhangPersonalizedFederatedInstruction2024}, and mixture of Feed-Forward Networks (FFN)~\cite{meiFedMoEPersonalizedFederated2024}, but they either (1) support only parameter-level personalization~\cite{qiFDLoRAPersonalizedFederated2024}, which may struggle with non-IID data; (2) rely on manually defined model architectures~\cite{zhangPersonalizedFederatedInstruction2024}, which may not align well with client data; or (3) incur significant communication costs from transmitting large-scale parameters~\cite{meiFedMoEPersonalizedFederated2024}.

\emph{Federated LLM Fine-Tuning with Mixture of Experts}
The Mixture-of-Experts (MoE) architecture~\cite{jacobsAdaptiveMixturesLocal1991,fedusSwitchTransformersScaling2021} offers a promising basis for PFL by enabling expert-group customization. Recent MoE-based PFL methods adopt Mixture of LoRA Experts (MoLE) to improve fine-tuning efficiency~\cite{wang2025adaptive}, but only support parameter personalization. Efforts on architectural heterogeneity either rely on dense FFN experts~\cite{meiFedMoEPersonalizedFederated2024}, which incur high communication costs and incompatibility with non-MoE LLMs like LLaMA~\cite{touvronLLaMAOpenEfficient2023}, or adopt manually defined model architectures~\cite{fanDeviceCollaborativeLanguage2024}, limiting adaptation to local data. In contrast, our method enables data-driven architectures, enhancing personalization while maintaining low communication cost.

\section{Problem Formulation}

We provide a formal definition of PFL for LLM fine-tuning.
In FL, $C$ clients collaboratively solve:
\begin{equation}
    \min_{\mathbf{\theta}}\mathcal{L}=\frac{1}{C}\sum_{i=1}^C\mathbb{E}_{\mathbf{X}\sim\mathcal{D}_i}\mathcal{L}_i(\mathbf{X}|\mathbf{\theta}),
\end{equation}
aiming at optimizing a global model $\mathbf{\theta}$ by minimizing the expected loss $\mathcal{L}_i$ over local data distribution $\mathcal{D}_i$ for each client $i$.
To address data heterogeneity across clients, PFL trains a personalized model $\mathbf{\theta}_i$ for each client $i$, modifying the objective to:
\begin{equation}
\min_{\mathbf{\theta}_1,\mathbf{\theta}_2...,\mathbf{\theta}_C}\mathcal{L}=\frac1C\sum_{i=1}^C\mathbb{E}_{\mathbf{X}\sim\mathcal{D}_i}\mathcal{L}_i(\mathbf{X}|\mathbf{\theta}_i).
\end{equation}

For LLM fine-tuning, $\mathbf{X}=[\mathbf{X}^{\text{instr}},\mathbf{X}^{\text{resp}}]$ denotes a text sequence, where $\mathbf{X}^{\text{instr}}$ is the instruction and $\mathbf{X}^{\text{resp}}$ is the response. $\mathcal{L}_i$ is the negative log-likelihood (NLL) loss defined as:
\begin{equation}
    \mathcal{L}_i(\mathbf{X} \mid \mathbf{\theta}_i)=-\frac{1}{T}\sum_{t=1}^{T}\text{log}\bigl[p(\mathbf{X}^{\text{resp}}_t \mid \mathbf{X}^{\text{instr}},\mathbf{X}^{\text{resp}}_{<t},\mathbf{\theta}_i)\bigr],
    \label{eq:nll_loss}
\end{equation}
where $T$ is the response sequence length.
In typical PFL, $\theta_i$ uses a uniform structure with varying parameters, while FedAMoLE enables architectural heterogeneity for greater personalization.
\section{Approach}

\subsection{Framework Overview}
\begin{figure*}[t]
    \centering
    \includegraphics[width=\textwidth]{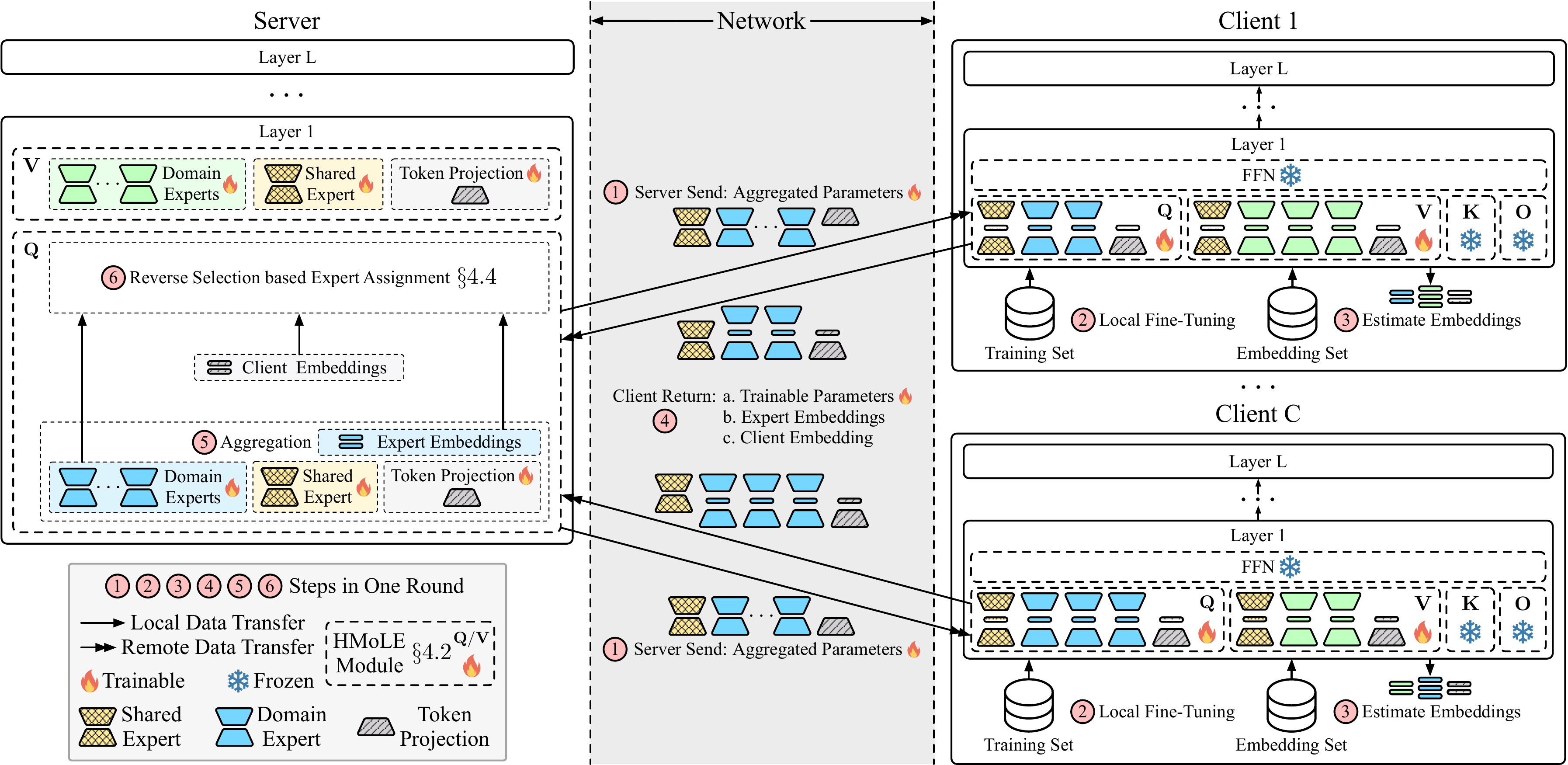}
    \caption{Overview of FedAMoLE in a single round. Each client has a transformer-based local model with $L$ decoder layers, where each layer includes a self-attention block (with parameters $\mathbf{Q}$, $\mathbf{K}$, $\mathbf{V}$, and $\mathbf{O}$) and an FFN block. Trainable HMoLE modules (see Figure~\ref{fig:moe_module}) are injected into $\mathbf{Q}$ and $\mathbf{V}$ for fine-tuning, while $\mathbf{K}$, $\mathbf{O}$, and FFN remain frozen. Step \ding{197} denotes the RSEA strategy (see Figure~\ref{fig:rsea_strategy}). Components of the same type (e.g., all shared experts) are shown with an identical color and texture.}
    \label{fig:workflow}
\end{figure*}
Figure~\ref{fig:workflow} overviews the components and processes of FedAMoLE. Each client has a local model with a frozen LLM backbone and trainable \emph{HMoLE modules}. 
Clients inject distinct HMoLE modules into corresponding linear layers, which allows architectural heterogeneity even with a common backbone.
The architecture of each HMoLE module is determined in a data-driven manner: 
for each activated linear layer, the server maintains a global HMoLE module comprising a shared expert, a pool of domain experts, and a token projection. 
The shared expert and token projection are broadcast to all clients, while client-specific subsets of domain experts are assigned with \emph{RSEA strategy} based on domain similarity, to form personalized HMoLE modules.
This design tailors personalized model architectures to both intra- (via domain experts) and inter-client (via shared experts) data distributions, and maximizes architectural flexibility through module-level expert assignment.

In FedAMoLE, each FL round consists of six steps.
First, the server distributes global parameters to all clients, including shared experts, domain experts, and token projections (Figure~\ref{fig:workflow}, \ding{192}).
Next, each client constructs HMoLE modules using these parameters and injects them into the shared LLM backbone to form its personalized model, which is then fine-tuned on its local training set (Figure~\ref{fig:workflow}, \ding{193}).
Then, each client computes embeddings for its domain experts and local data using a randomly sampled subset of its training set, termed \emph{embedding set} (Figure~\ref{fig:workflow}, \ding{194}), and returns the fine-tuned parameters along with these embeddings to the server (Figure~\ref{fig:workflow}, \ding{195}).
Finally, the server aggregates the received parameters and embeddings (Figure~\ref{fig:workflow}, \ding{196}), and applies the RSEA strategy to determine the model architecture for each client in the next round (Figure~\ref{fig:workflow}, \ding{197}).

\subsection{Heterogeneous MoLE Module} \label{sec:hmole_module}
\begin{figure}[t]
    \centering
    \includegraphics[width=0.9\linewidth]{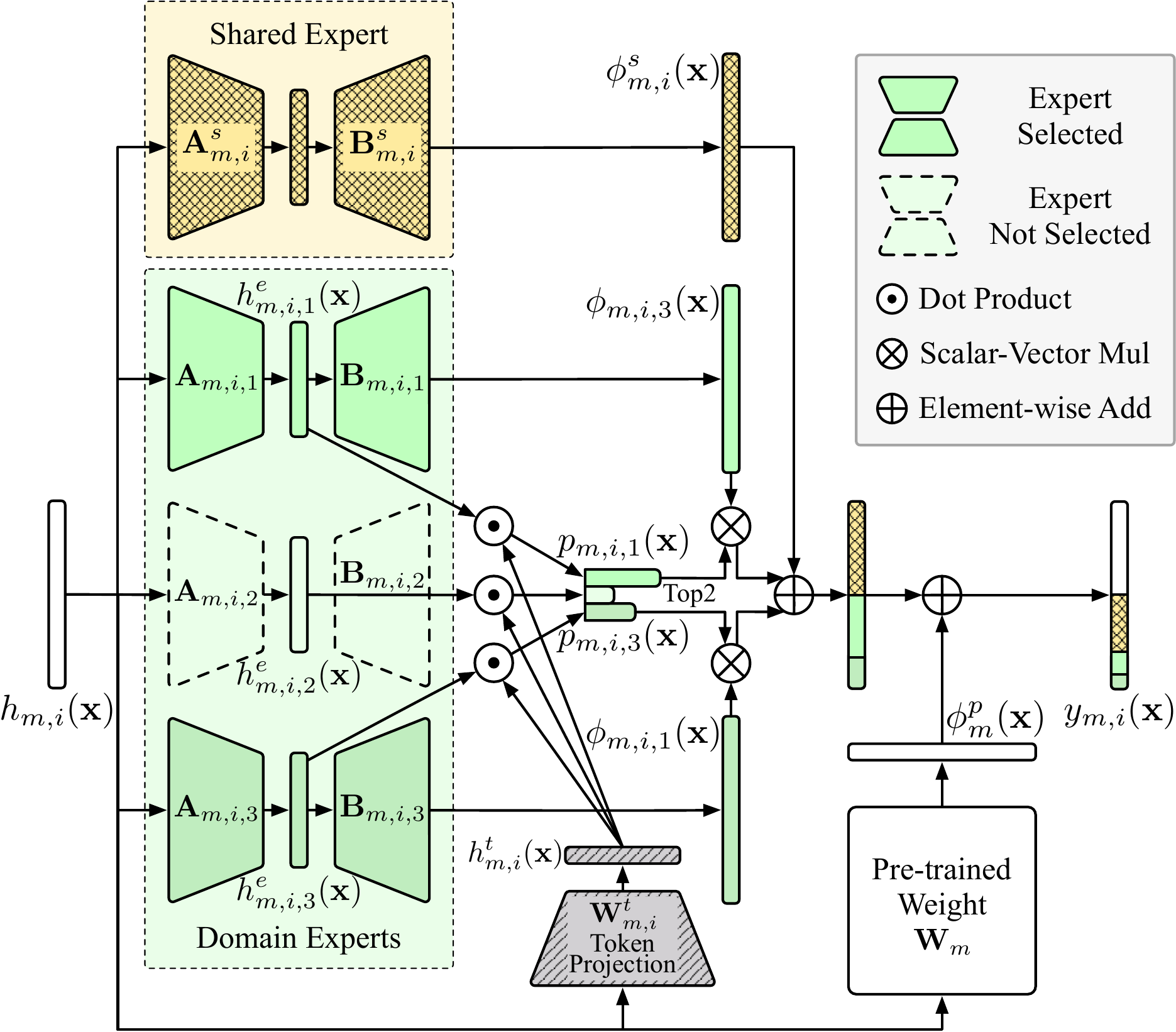}
    \caption{Routing of HMoLE module $m$ at client $i$.}
    \label{fig:moe_module}
\end{figure}
While vanilla MoLE enables personalized architectures in a lightweight manner, it poses aggregation challenges in FL due to expert-number-dependent router shapes (see \S\ref{sec:preliminaries}). To address this, we propose the HMoLE module, which decouples router shape from local expert configuration, thereby enabling lightweight model heterogeneity with FedAvg compatibility.

The routing process of HMoLE is shown in Figure~\ref{fig:moe_module}. 
Let $h_{m,i}(\mathbf{x}) \in \mathbb{R}^d$ denote the hidden state of input token $\mathbf{x}$ in module $m$ of client $i$.
It is projected into a low-dimensional subspace by matrix $\mathbf{A}$ of each domain expert $j$, generating the corresponding expert embedding $h^e_{m,i,j}(\mathbf{x}) \in \mathbb{R}^r$. Also, $h_{m,i}(\mathbf{x})$ is projected by the token projection matrix $\mathbf{W}^t_{m,i} \in \mathbb{R}^{r \times d}$ to obtain the token embedding $h^t_{m,i}(\mathbf{x}) \in \mathbb{R}^r$ in this subspace. Next, the scaled dot product between the token embedding and each expert embedding is computed, followed by softmax normalization to obtain the weighting coefficient $p_{m,i,j}(\mathbf{x})$ for each domain expert $j$. Finally, $k^e$ domain experts with the highest coefficients (where $k^e=2$ in Figure~\ref{fig:moe_module}) are selected to extract features $\phi_{m,i,j}(\mathbf{x})$. These features are averaged based on the corresponding coefficients, and combined with the feature from the shared expert $\phi^s_{m,i}(\mathbf{x})$ and the pre-trained feature $\phi^p_m(\mathbf{x})$ to get the final output $y_{m,i}(\mathbf{x})$. This process can be formalized as
\allowdisplaybreaks
\begin{align}
    &h^e_{m,i,j}(\mathbf{x})=\mathbf{A}_{m,i,j} h_{m,i}(\mathbf{x}) \label{eq:expert_embedding} \\
    &h^t_{m,i}(\mathbf{x})=\mathbf{W}^t_{m,i} h_{m,i}(\mathbf{x}) \label{eq:token_embedding} \\
    &p_{m,i,j}(\mathbf{x})=\text{softmax}\left[\frac{{h^t_{m,i}(\mathbf{x})}^T h^e_{m,i,j}(\mathbf{x})}{\sqrt{d}}\right] \label{eq:routing_prob} \\
    &\mathcal{T}_{m,i}(\mathbf{x})=\text{top-}k^e(p_{m,i,j}(\mathbf{x})) \\
    &\phi^p_m(\mathbf{x})=\mathbf{W}_{m} h_{m,i}(\mathbf{x}) \\
    &\phi^s_{m,i}(\mathbf{x})=\mathbf{B}^s_{m,i} \mathbf{A}^s_{m,i} h_{m,i}(\mathbf{x}) \\
    &\phi_{m,i,j}(\mathbf{x})=\mathbf{B}_{m,i,j} \mathbf{A}_{m,i,j} h_{m,i}(\mathbf{x}) \\
    &y_{m,i}(\mathbf{x})=\phi^p_m(\mathbf{x}) + \phi^s_{m,i}(\mathbf{x}) + \sum_{j\in\mathcal{T}_{m,i}(\mathbf{x})} p_{m,i,j}(\mathbf{x}) \cdot \phi_{m,i,j}(\mathbf{x}),
\end{align}
where $\mathbf{W}_m$ is the pre-trained weight, $\mathbf{A}_{m,i,j}$ and $\mathbf{B}_{m,i,j}$ are the parameters of domain expert $j$, $\mathbf{A}^s_{m,i}$ and $\mathbf{B}^s_{m,i}$ are those of the shared expert, and $k^e$ is the number of domain experts selected per token.

\subsection{Local Fine-Tuning and Aggregation}
In each round, client $i$ freezes the pre-trained backbone and trains the parameters of its HMoLE modules $\mathcal{M}$, i.e., $\mathbf{\theta}_i = \bigcup_{m \in \mathcal{M}} \mathbf{\theta}_{m,i}$, where $\mathbf{\theta}_{m,i} = \{ \mathbf{A}^s_{m,i}, \mathbf{B}^s_{m,i}, \mathbf{W}^r_{m,i} \} \cup \{ \mathbf{A}_{m,i,j}, \mathbf{B}_{m,i,j} \}_{j \in \mathcal{E}_{m,i}}$ is the trainable parameters of module $m$ on client $i$, and $\mathcal{E}_{m,i}$ is the set of domain experts assigned to this module.
In each iteration, client $i$ samples a mini-batch $\mathcal{B}$ from its training set $\mathcal{D}^{\text{train}}_i$ and computes the gradient of 
its parameters $\mathbf{\theta}_i$ w.r.t. the loss function
\begin{equation}
    \mathcal{L}_i (\mathcal{B} \mid \mathbf{\theta}_i) = \frac{1}{|\mathcal{B}|} \sum_{\mathbf{X} \in \mathcal{B}} \left[ \mathcal{L}_i (\mathbf{X} \mid \mathbf{\theta}_i) + \beta\cdot\mathcal{L}^{\text{LB}}_i(\mathcal{B} \mid \mathbf{\theta}_i) \right],
\end{equation}
where $\mathcal{B}$ is a mini-batch of text sequences containing $T$ tokens sampled from client $i$'s training dataset $\mathcal{D}^{\text{train}}_i$, $\mathcal{L}_i$ represents the NLL loss (refer to (\ref{eq:nll_loss})), and
\begin{equation}
\mathcal{L}^{\text{LB}}_i(\mathcal{B}\mid\mathbf{\theta}_i)=\sum_{m\in\mathcal{M}}\left(\left|\mathcal{E}_{m,i}\right|\sum_{j\in\mathcal{E}_{m,i}}\overline{f}_{m,i,j}\cdot \overline{p}_{m,i,j}\right)
\end{equation}
represents the load balance loss~\cite{fedusSwitchTransformersScaling2021} weighted by $\beta$, which encourages each domain expert to process a comparable number of tokens, ensuring sufficient training. $\overline{f}_{m,i,j}$ and $\overline{p}_{m,i,j}$ are defined as
\begin{equation}
    \begin{aligned}
        &\overline{f}_{m,i,j}=\frac{1}{T}\sum_{\mathbf{X}\in\mathcal{B}}\sum_{\mathbf{x}\in\mathbf{X}}\mathbbm{1}\{\mathop{\arg\max}\limits_{j'}p_{m,i,j'}(\mathbf{x})=j\}\\
        &\overline{p}_{m,i,j}=\frac{1}{T}\sum_{\mathbf{X}\in\mathcal{B}}\sum_{\mathbf{x}\in\mathbf{X}}p_{m,i,j}(\mathbf{x}).
    \end{aligned}
\end{equation}
The parameters are then updated by $\mathbf{\theta}_i^{t+1} = \mathbf{\theta}_i^t - \eta\nabla_{\mathbf{\theta}_i^t}\mathcal{L}_i (\mathcal{B} | \mathbf{\theta}_i^t)$, where $\eta$ is the learning rate.
After fine-tuning, each HMoLE module $m$ performs federated aggregation as
\allowdisplaybreaks
\begin{equation}
    \mathbf{X}_{m,j}=\frac{1}{\left|\mathcal{C}_{m,j}\right|}\sum_{i\in\mathcal{C}_{m,j}}\mathbf{X}_{m,i,j}, \quad \mathbf{Y}_{m}=\frac{1}{C}\sum_{i=1}^C\mathbf{Y}_{m,i},
\end{equation}
where $\mathcal{C}_{m,j}$ denotes the set of clients participating in fine-tuning domain expert $j$ within module $m$ in this round, $\mathbf{X}$ is applied to both $\mathbf{A}$ and $\mathbf{B}$, $\mathbf{Y}$ is applied to both $\mathbf{A}^s$, $\mathbf{B}^s$ and $\mathbf{W}^t$.
Owing to the small scale of HMoLE modules (around 1\% of the total parameters), memory overhead of local fine-tuning and the communication cost of aggregation remain minimal (Tables~\ref{tab:system_overhead} and~\ref{tab:switch_transformer_overhead}).

\subsection{Expert Assignment with Reverse Selection}
\begin{figure}[t]
    \centering
    \includegraphics[width=0.9\linewidth]{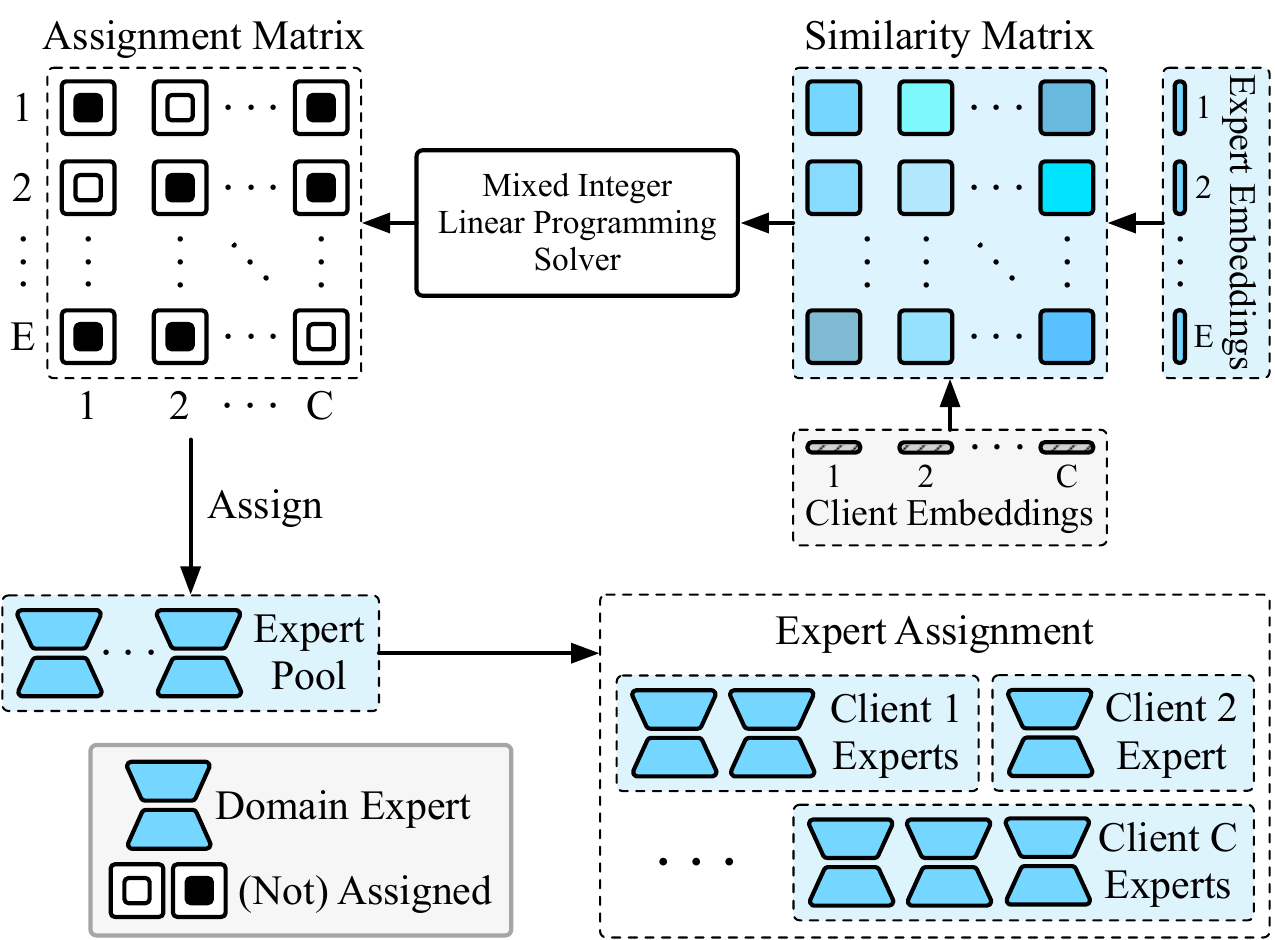}
    \caption{RSEA process. $E$ and $C$ denote the number of domain experts and clients, respectively.}
    \vspace{-0.15cm}
    \label{fig:rsea_strategy}
\end{figure}
To enable data-aware model architectures in MoE-based FL, traditional approaches~\cite{dunFedJETsEfficientJusttime2023} adopt a forward selection strategy: each client first downloads all experts from the server and evaluates their relevance to its local data using a predefined metric. Then, \emph{each client selects the $k$ most relevant experts} to form its local model. However, this strategy (1) enforces an identical model structure, as all clients select the same number of experts; (2) may lead to suboptimal expert assignment, as it considers only local data and ignores the collaborative nature of expert training; (3) risks privacy leakage, as clients may infer peers' data characteristics by accessing all experts; and (4) incurs high communication cost from downloading the full expert pool.

To address these limitations, we propose a \textbf{r}everse \textbf{s}election-based \textbf{e}xpert \textbf{a}ssignment (RSEA) strategy (see Figures~\ref{fig:workflow} and ~\ref{fig:rsea_strategy}): each client first uploads embeddings of its local data and assigned experts. On the server side, \emph{each expert then selects the $k$ most relevant clients} based on an embedding similarity metric. Finally, each client downloads only the experts that have selected it to construct the local model. Compared to forward selection, RSEA (1) enables model heterogeneity by allowing each client to receive a different number of experts; (2) approximates optimal expert assignment by considering data characteristics from all clients; (3) enhances privacy by restricting access to only relevant experts; and (4) reduces communication costs by limiting downloads to a subset of experts.

RSEA strategy requires a metric to quantify the relevance between client data and expert domains. Thus, we define the embedding of client $i$ on module $m$ as the element-wise mean of token embeddings in $m$ for all tokens in the embedding set $\mathcal{D}^{\text{emb}}_i$ sampled from the training set $\mathcal{D}_i^{\text{train}}$:
\begin{equation}
    \overline{\mathbf{h}}^t_{m,i}=\frac{1}{\left|\mathcal{D}^{\text{emb}}_i\right|}\sum_{\mathbf{x}\in\mathcal{D}^{\text{emb}}_i}h^t_{m,i}(\mathbf{x}). \label{eq:client_embedding}
\end{equation}
Note that the sampling of $\mathcal{D}^{\text{emb}}_i$ is random to ensure its data distribution aligns with $\mathcal{D}^{\textbf{train}}_i$, thereby better reflecting the local data characteristics.
Similarly, we define the embedding of domain expert $j$ in module $m$ as the element-wise mean of $j$'s embeddings across clients $\mathcal{C}_{m,j}$ that fine-tune $j$ in the current round (\ref{eq:expert_embedding_aggr}), where $\overline{\mathbf{h}}^e_{m,i,j}$ is the embedding of $j$ on $i$, i.e., the element-wise mean of $j$'s embeddings in $m$ for all tokens in $\mathcal{D}^{\text{emb}}_i$ (\ref{eq:expert_embedding_single}).
\begin{align}
    &\overline{\mathbf{h}}^e_{m,j}=\frac{1}{\left|\mathcal{C}_{m,j}\right|}\sum_{i\in\mathcal{C}_{m,j}}\overline{\mathbf{h}}^e_{m,i,j} \label{eq:expert_embedding_aggr} \\
    &\overline{\mathbf{h}}^e_{m,i,j}=\frac{1}{\left|\mathcal{D}^{\text{emb}}_i\right|}\sum_{\mathbf{x}\in\mathcal{D}^{\text{emb}}_i}h^e_{m,i,j}(\mathbf{x}) \label{eq:expert_embedding_single}
\end{align}

The metric $s_{m,i,j}$, measuring the relevance between client $i$'s data and expert $j$'s domain within module $m$, is then defined as the scaled dot product of their embeddings (\ref{eq:hmole_relevance}), which aligns with the HMoLE training objective (\ref{eq:expert_embedding}–\ref{eq:routing_prob}).
A higher $s_{m,i,j}$ signifies a stronger relevance, indicating a greater likelihood of expert $j$ being selected by client $i$'s tokens in module $m$.
This design establishes a linkage between experts and the hidden domains in client data, which gradually stabilizes during training and ultimately aligns each expert with the most relevant clients~(see \S\ref{sec:discuss_expert_assign}).
\begin{equation}
    s_{m,i,j}=\frac{\left(\overline{\mathbf{h}}^t_{m,i}\right)^T\cdot \overline{\mathbf{h}}^e_{m,j}}{\sqrt{d}} \label{eq:hmole_relevance}
\end{equation}

With this metric, each domain expert $j$ in module $m$ selects $k^c$ clients with the highest relevance:
\begin{equation}
    \mathcal{C}_{m,j}=\text{top-}k^c(\text{softmax}(s_{m,i,j})).
    \label{eq:reverse_selection}
\end{equation}

Note that we avoid using a threshold-based approach here, as (1) a continuous threshold introduces a much larger search space than discrete client counts, making hyperparameter tuning more difficult, and (2) thresholding may waste model capacity, e.g., if the threshold is 0.4 and the relevance scores of an expert to three clients are all 1/3, the expert would not be assigned to any client.

However, the reverse selection strategy (\ref{eq:reverse_selection}) may lead to imbalanced expert assignments, where some clients are selected by no experts while others are selected by all. To constrain the minimum number of experts per module ($k^e$) and limit expert heterogeneity (up to $b$ experts per module per client), we formulate expert assignment within each module as the following optimization problem:

\vspace{-0.15cm}
\begin{definition}(Domain Expert Assignment Problem)
    \begin{align*}
        &\max_{\mathbf{D}_m} \left\langle \mathbf{P}_m, \mathbf{D}_m \right\rangle \\
        \text{s.t.}\quad&\forall i: k^e\leq\sum_{j} d_{m,i,j}\leq b \\
        &\forall j: \sum_{i} d_{m,i,j}=k^c \\
        &\forall i,j: d_{m,i,j}\in\{0,1\}.
    \end{align*}
    \vspace{-0.56cm}
    \label{def:expert_assign_opt}
\end{definition}
Here, $\mathbf{P}_m = \{ p_{m,i,j} \}$ is the probability matrix for selection, where $p_{m,i,j} = \mathrm{e}^{s_{m,i,j}} / \sum_i \mathrm{e}^{s_{m,i,j}}$ denotes the probability that domain expert $j$ in module $m$ selects client $i$. $\mathbf{D}_m = \{ d_{m,i,j} \}$ is the binary assignment matrix, with $d_{m,i,j} = 1$ if expert $j$ is assigned to client $i$, and $0$ otherwise. The objective maximizes the inner product $\left\langle \mathbf{P}_m, \mathbf{D}_m \right\rangle$, aligning expert assignments with selection preferences. The first constraint bounds the number of experts assigned to each client between $k^e$ and $b$ for balance, while the second constraint enforces that each expert selects exactly $k^c$ clients, as in (\ref{eq:reverse_selection}).

Problem~\ref{def:expert_assign_opt} can be reduced to a mixed integer linear programming (MILP) problem, which is $\mathcal{NP}$-hard. However, for domain expert assignment, the problem size is typically small (a few hundred decision variables) and can be quickly solved using solvers such as Gurobi~\cite{gurobi} or SCIP~\cite{BolusaniEtal2024OO}.

\subsection{Enhancement on Privacy} \label{sec:dp_method}
RSEA strategy requires each client $i$ to upload its expert embeddings $\{\overline{\mathbf{h}}^e_{m,i,j}\}_{j\in\mathcal{E}_{m,i}}$ and data embedding $\overline{\mathbf{h}}^t_{m,i}$.
While these embeddings do not directly reveal raw data, techniques such as differential privacy (DP) \cite{wei2020federated} can be employed to provide stronger and more comprehensive privacy protection.
For example, to privatize $\overline{\mathbf{h}}^e_{m,i,j}$, we can adopt the DP-based Split-N-Denoise method~\cite{maiSplitanddenoiseProtectLarge2024}, which adds Laplace noise $\mathbf{z} \sim c\cdot\exp(-\eta\cot\left\|\mathbf{z}\right\|_2)$ (with constants $c$ and $\eta$) to obtain $M(\overline{\mathbf{h}}_{m,i,j}^e) = \overline{\mathbf{h}}_{m,i,j}^e + \mathbf{z}$. The result is then $l_2$-clipped as $M^{\prime}(\overline{\mathbf{h}}_{m,i,j}^e) = M(\overline{\mathbf{h}}_{m,i,j}^e) \cdot \min(1, C/||M(\overline{\mathbf{h}}_{m,i,j}^e)||_2)$, where $C$ is the upper bound of $||\overline{\mathbf{h}}_{m,i,j}^e||_2$. Finally, $M^{\prime}(\overline{\mathbf{h}}_{m,i,j}^e)$ satisfies $\eta d_\chi$-privacy and can be safely uploaded. We further show in experiments that FedAMoLE maintains strong performance with DP (see \S\ref{sec:dp_exp}).
\section{Experiments}
\begin{table*}[t]
    \caption{
    Comparison of fine-tuning accuracy (MTAL) across seven scenarios. Each cell reports the mean and standard deviation of MTAL calculated over three random seeds. Each column represents the results under a specific data heterogeneity scenario, where the best-performing approach is highlighted in \textbf{bold}, the second-best method is \underline{underlined}, ``gains'' denote the relative MTAL improvement of FedAMoLE over the best baseline, and ``-'' indicates failure to converge in the official implementation.
    }
    \setlength\tabcolsep{3.8pt}
    \centering
    \begin{tabular}{c|l|c|c|c|ccc|c|c}
        \toprule
        \multicolumn{2}{c|}{\multirow{2}{*}{Approach}} & \textbf{BigBench} & \textbf{Hybrid} & \textbf{NI} & \multicolumn{3}{c|}{\textbf{Dolly-15K}} & \textbf{SNLI} & \multirow{2}{*}{Average} \\
        \multicolumn{2}{c|}{} & 1 domain/client & 1 task/client & 1 task/client & $\alpha=0.1$ & $\alpha=1.0$ & $\alpha=100.0$ & $\alpha=1.0$ & \\
        \midrule
        \multirow{3}{*}{\rotatebox{90}{Vanilla}} & FedIT & 43.84$\pm$1.48 & 59.26$\pm$0.72 & 54.09$\pm$0.23 & 28.69$\pm$1.73 & 28.67$\pm$1.28 & \underline{27.31$\pm$0.55} & 83.06$\pm$0.63 & 46.42 \\
        & FedPrompt & 26.80$\pm$1.95 & 39.93$\pm$1.51 & 35.45$\pm$1.08 & 26.01$\pm$2.16 & 26.44$\pm$1.22 & 24.57$\pm$0.78 & 66.10$\pm$2.78 & 35.04 \\
        & FedPTuning & 42.45$\pm$0.79 & 50.92$\pm$11.78 & 36.10$\pm$18.36 & 24.11$\pm$3.14 & 27.57$\pm$0.79 & 26.16$\pm$1.12 & 76.83$\pm$6.14 & 40.59 \\
        \midrule
        \multirow{7.5}{*}{\rotatebox{90}{Personalized}} & FedIT-FT & 47.67$\pm$1.18 & 60.17$\pm$0.90 & 53.34$\pm$0.26 & \underline{28.80$\pm$0.16} & \underline{29.13$\pm$1.18} & 27.17$\pm$0.40 & \underline{86.67$\pm$0.58} & \underline{47.56} \\
        & FedPrompt-FT & 26.68$\pm$1.05 & 40.87$\pm$2.06 & 47.70$\pm$0.67 & 26.76$\pm$0.57 & 25.91$\pm$0.98 & 24.63$\pm$1.17 & 75.75$\pm$1.46 & 38.33 \\
        & FedPTuning-FT & 43.60$\pm$1.36 & 52.39$\pm$12.56 & 38.97$\pm$20.72 & 27.28$\pm$1.28 & 28.05$\pm$1.42 & 25.89$\pm$0.43 & 86.34$\pm$3.20 & 43.22 \\
        & CoMiGS & 29.40$\pm$0.80 & 35.25$\pm$1.44 & 25.26$\pm$0.67 & - & - & - & 62.09$\pm$2.40 & 38.00 \\
        & FDLoRA & \underline{48.23$\pm$1.85} & \underline{60.31$\pm$0.95} & \underline{57.04$\pm$1.90} & 27.75$\pm$0.17 & 26.67$\pm$0.66 & 25.79$\pm$1.99 & 86.06$\pm$1.32 & 47.41 \\
        \cmidrule(lr){2-10}
        & FedAMoLE & \textbf{52.56$\pm$1.43} & \textbf{63.60$\pm$0.30} & \textbf{60.04$\pm$1.09} & \textbf{29.87$\pm$1.40} & \textbf{29.72$\pm$1.43} & \textbf{28.28$\pm$0.84} & \textbf{88.78$\pm$0.38} & \textbf{50.41} \\
        & Gains (\%) & 8.97 & 5.46 & 5.25 & 3.72 & 2.02 & 3.55 & 2.43 & 5.97 \\
        \bottomrule
    \end{tabular}
    \label{tab:compare_acc}
\end{table*}
\begin{figure*}[t]
    \vspace{0.15cm}
    \centering
    \begin{subfigure}[c]{0.245\textwidth}
        \centering
        \includegraphics[width=\linewidth]{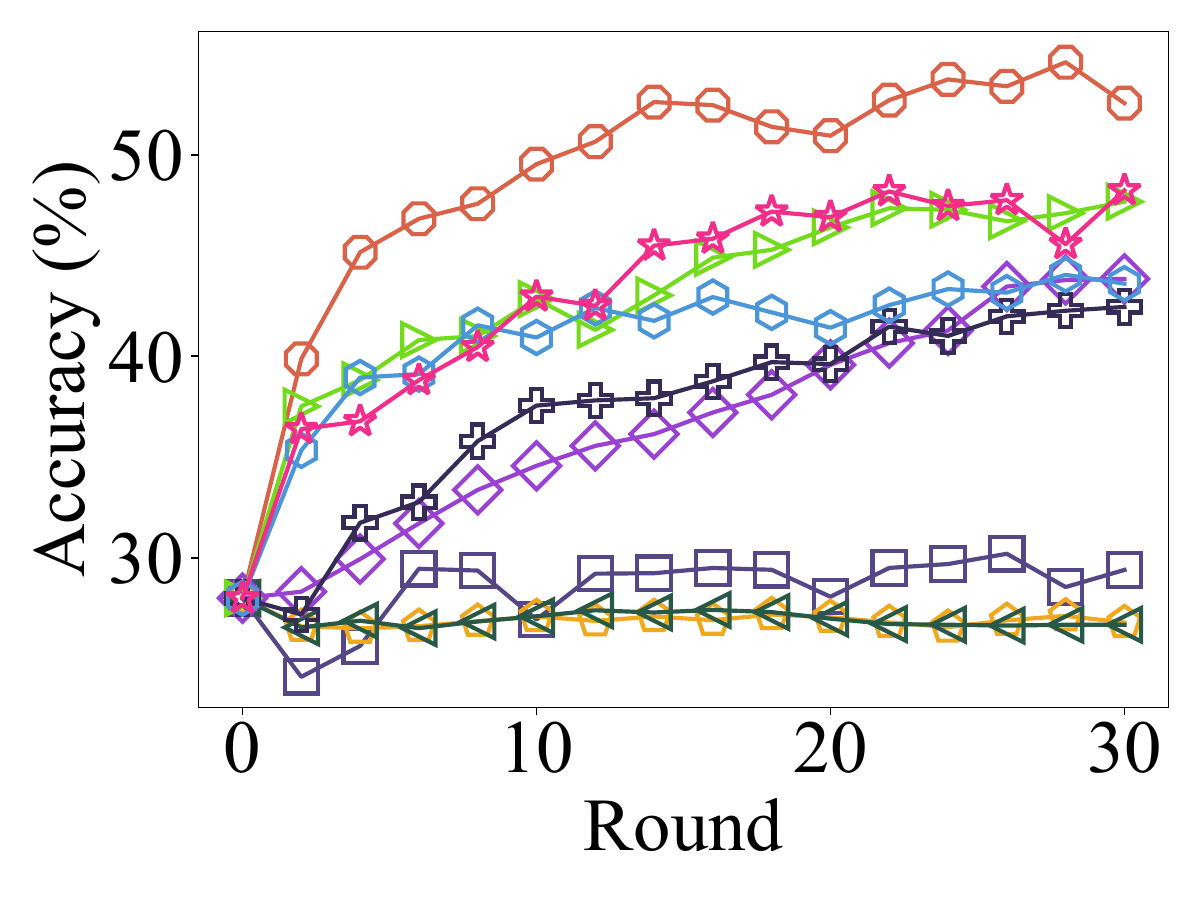}
        \caption{BigBench 1 domain/client}
        \label{fig:bigbench_meta1}
    \end{subfigure}
    \begin{subfigure}[c]{0.245\textwidth}
        \centering
        \includegraphics[width=\linewidth]{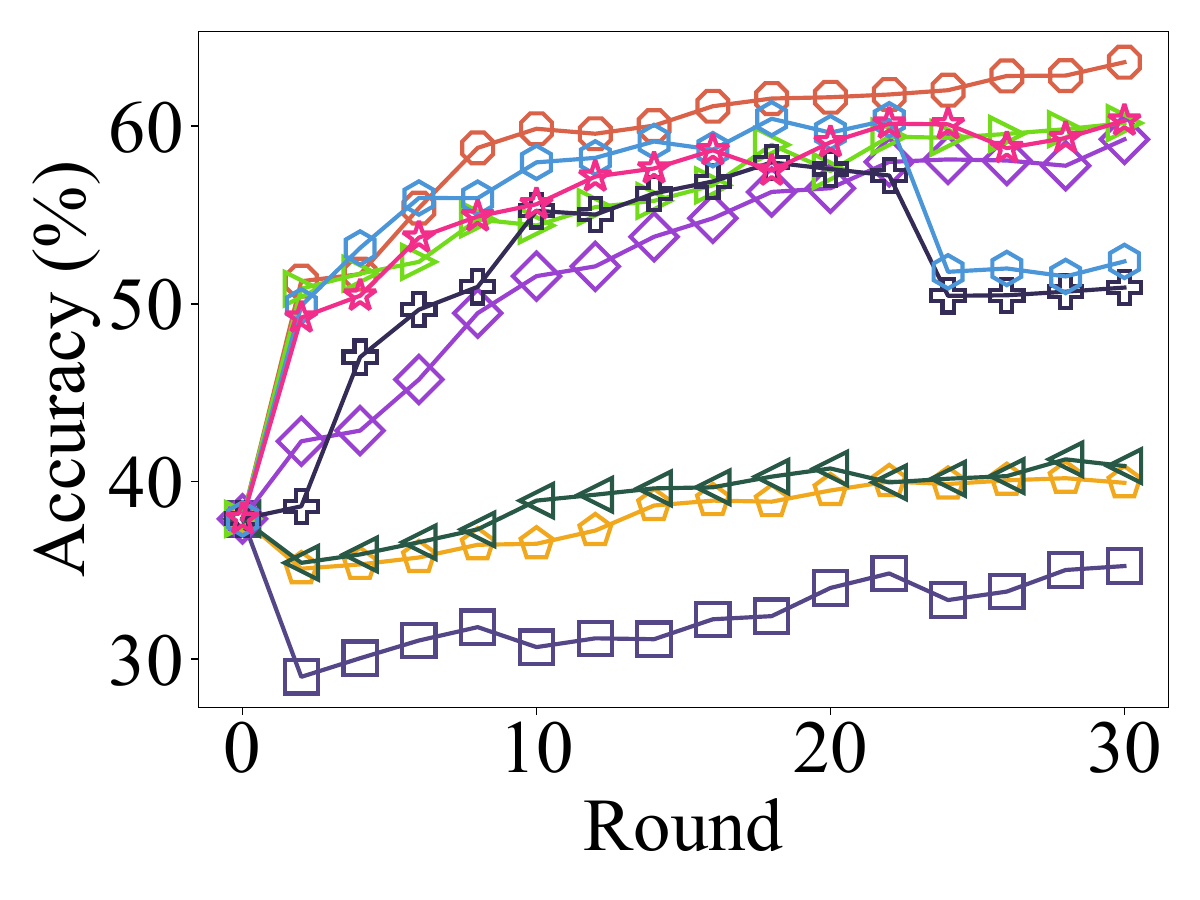}
        \caption{Hybrid 1 task/client}
        \label{fig:hybrid_meta1}
    \end{subfigure}
    \begin{subfigure}[c]{0.245\textwidth}
        \centering
        \includegraphics[width=\linewidth]{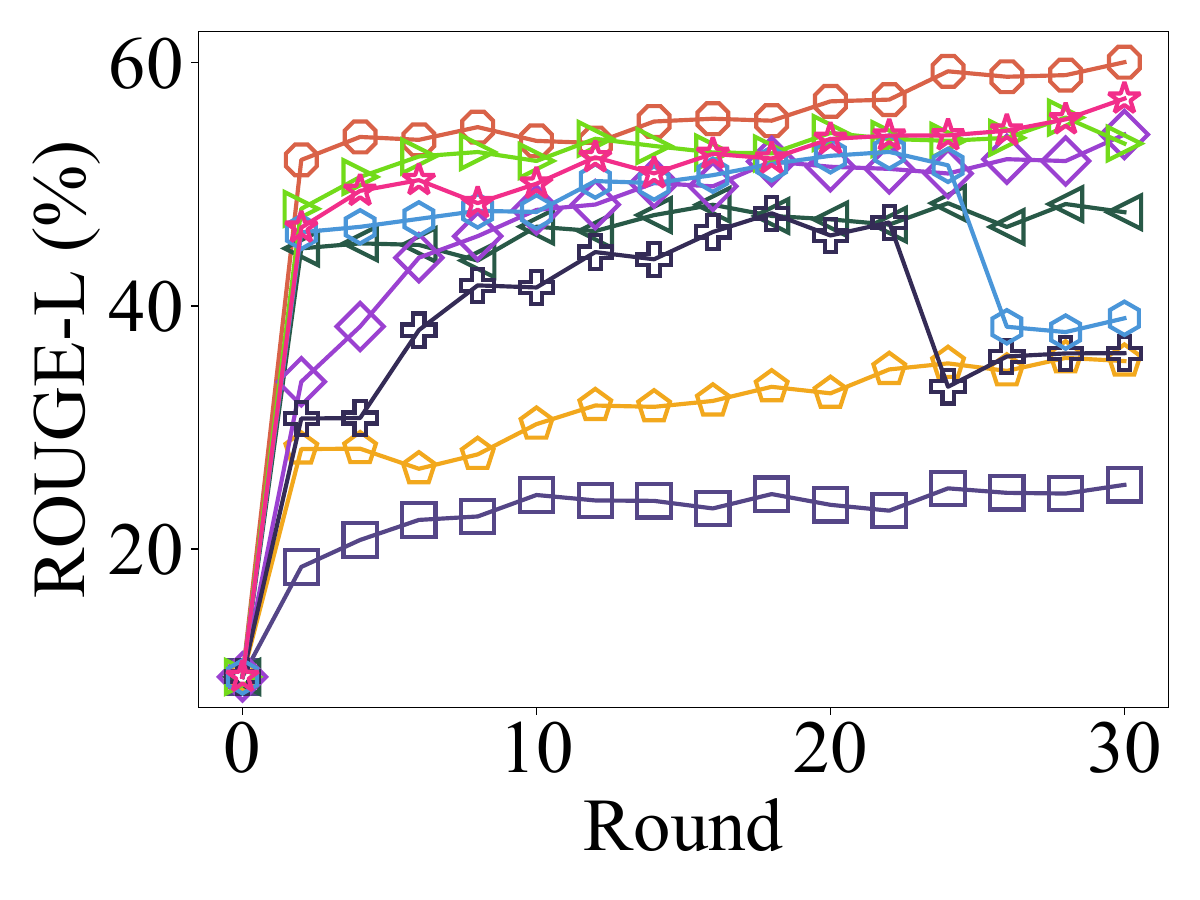}
        \caption{NI 1 task/client}
        \label{fig:natural_instruct_meta1}
    \end{subfigure}
    \begin{subfigure}[c]{0.245\textwidth}
        \centering
        \includegraphics[width=\linewidth]{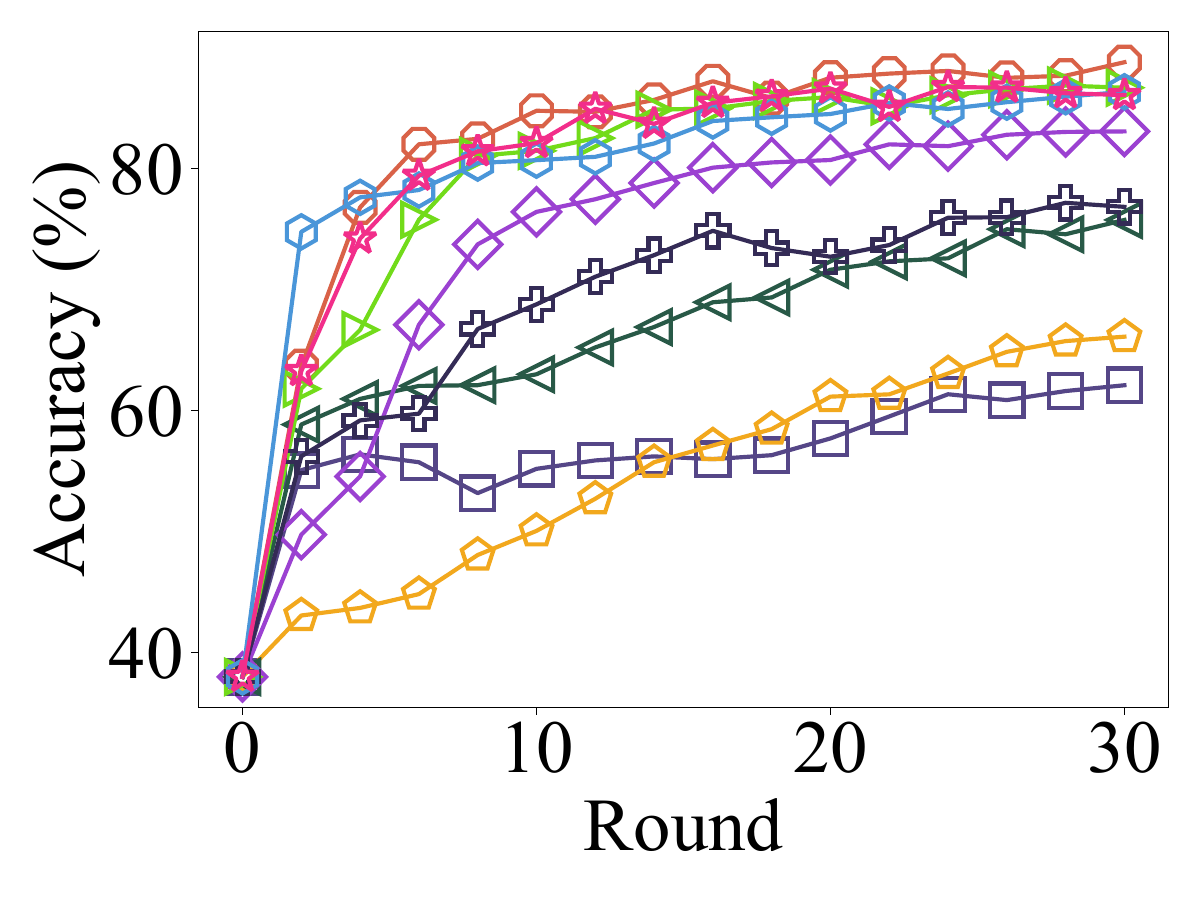}
        \caption{SNLI $\alpha=1.0$}
        \label{fig:snli_dir1.0}
    \end{subfigure}
    \begin{subfigure}[c]{\textwidth}
        \vspace{0.5em}
        \centering
        \includegraphics[width=\linewidth]{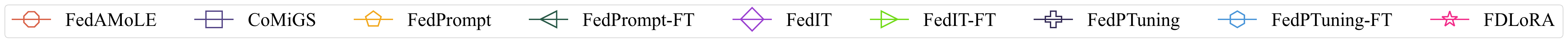}
        \caption*{}
    \end{subfigure}
    \vspace{-0.8cm}
    \caption{Trends of MTA among comparison methods, averaged over three runs with different seeds.}
    \label{fig:acc_per_round}
    \vspace{0.04cm}
\end{figure*}

The experiments aim at demonstrating: 
1) FedAMoLE achieves better accuracy on non-IID data (\S\ref{sec:exp_noniid}), 
2) FedAMoLE has lower resource overhead than MoE-based approaches (\S\ref{sec:exp_system}), 
and 3) the components proposed in this work contribute positively (\S\ref{sec:ablation}-\ref{sec:discuss_expert_assign}).

\subsection{Experimental Setup} \label{sec:exp_setup}
\emph{Baselines.}
We compare FedAMoLE to 8 federated LLM fine-tuning methods, encompassing both non-personalized and personalized ones. Non-personalized ones include FedIT~\cite{zhangBuildingFederatedgptFederated2023}, FedPrompt, and FedPTuning~\cite{kuangFederatedScopeLLMComprehensivePackage2023}, which adopt FedAvg~\cite{mcmahan2017fl} with LoRA~\cite{hu2022lora}, Prompt Tuning~\cite{lesterPowerScaleParameterEfficient2021}, and P-Tuning~\cite{liuPtuningPromptTuning2022} respectively. Personalized ones include FedIT-FT, FedPrompt-FT, FedPTuning-FT (each tunes its trainable modules after aggregation~\cite{yuSalvagingFederatedLearning2022}), FDLoRA~\cite{qiFDLoRAPersonalizedFederated2024} which employs shared and personalized adapters combined via few-shot black-box optimization, and CoMiGS~\cite{fanDeviceCollaborativeLanguage2024} which uses manually defined personalized MoLE structures where clients alternately train local routers and experts, aggregating only the shared expert.

\emph{FL Settings.}
Unless otherwise stated, we construct a 10-client cross-silo FL scenario. All experiments use LLaMA-3.2-1B\footnote{Adopt checkpoint from \url{https://huggingface.co/meta-llama/Llama-3.2-1B}.} as the foundation model and run for 30 rounds, with each client performing 200 local updates per round.

\emph{Datasets.}
To comprehensively evaluate FL methods in realistic heterogeneous data scenarios, we construct three kinds of non-IID data heterogeneity. Aligning with scenarios of collaboration across organizations from distinct domains, we first introduce heterogeneity from domain skew by assigning each of the 10 clients a task from a distinct domain in BigBench~\cite{srivastava2022beyond}. For scenarios of collaboration among organizations holding different task types within the same or similar domains, we then create heterogeneity from task skew, where each client is assigned a unique task from Natural Instructions (NI)~\cite{wangSuperNaturalInstructionsGeneralizationDeclarative2022} or our custom Hybrid dataset (contains 10 subsets, detailed in Appendix~\ref{sec:impl}). Finally, we simulate heterogeneity via label skew following prior studies \cite{qin2023fedapen,kuangFederatedScopeLLMComprehensivePackage2023} by partitioning SNLI~\cite{bowmanLargeAnnotatedCorpus2015} and Dolly-15K~\cite{DatabricksBlog2023DollyV2} among clients using a Dirichlet distribution with $\alpha$ values of 0.1, 1.0, and 100.0.

\emph{Evaluation.}
Following prior studies on personalized federated fine-tuning \cite{yangDualpersonalizingAdapterFederated2024,qiFDLoRAPersonalizedFederated2024}, we construct an in-domain/task test set for each client and adopt \textbf{m}ean \textbf{t}est \textbf{a}ccuracy at the \textbf{l}ast round (MTAL) as the evaluation metric, defined as:
\begin{equation}
    \text{MTAL}=\text{MTA}[-1].
\end{equation}
Here, $\text{MTA}[t] = \frac{1}{C}\sum_{i=1}^C acc_i[t]$, where $acc_i[t]$ evaluates the performance of client $i$'s local model at round $t$. 
For classification tasks, $acc$ denotes accuracy, while for natural language generation tasks, $acc$ corresponds to ROUGE-L~\cite{linROUGEPackageAutomatic} following prior studies on federated LLM tuning ~\cite{qinFederatedFullParameterTuning2023b,kuangFederatedScopeLLMComprehensivePackage2023}.

\subsection{Comparison of Accuracy} \label{sec:exp_noniid}
From Table~\ref{tab:compare_acc}, FedAMoLE achieves the best results across all seven scenarios. Its advantages are most pronounced on the highly heterogeneous datasets—BigBench, Hybrid, and NI—where each client's data belongs to a unique domain or task. On these, FedAMoLE outperforms the strongest baseline, FDLoRA, by substantial margins of 8.97\% on BigBench, 5.46\% on Hybrid, and 5.25\% on NI. In contrast, CoMiGS, another MoLE-based method, struggles on these complex tasks, highlighting the superiority of FedAMoLE's design. This stems from three key advantages: 1) its data-driven architecture better adapts to local data than CoMiGS's manual setup; 2) its aggregatable router provides superior generalization over a local-only router; and 3) its domain experts are collaboratively trained, enhancing performance beyond single-client training. Furthermore, FedAMoLE's effectiveness extends to label-skewed scenarios, improving accuracy by 2.43\% on SNLI ($\alpha=1.0$) over FedIT-FT while also consistently achieving the best performance across all Dolly-15K settings. This consistent gain is due to FedAMoLE’s data-aware, module-level adaptive expert assignment strategy, which builds personalized models tailored to each client’s local data, enhancing performance on heterogeneous data (\S\ref{sec:discuss_expert_assign}).
We further evaluate FedAMoLE on additional LLMs to show its generalization in \S\ref{sec:compare_acc_extended}.
\begin{figure}[H]
    \centering
    \begin{subfigure}[t]{0.47\linewidth}
        \centering
        \includegraphics[width=\linewidth]{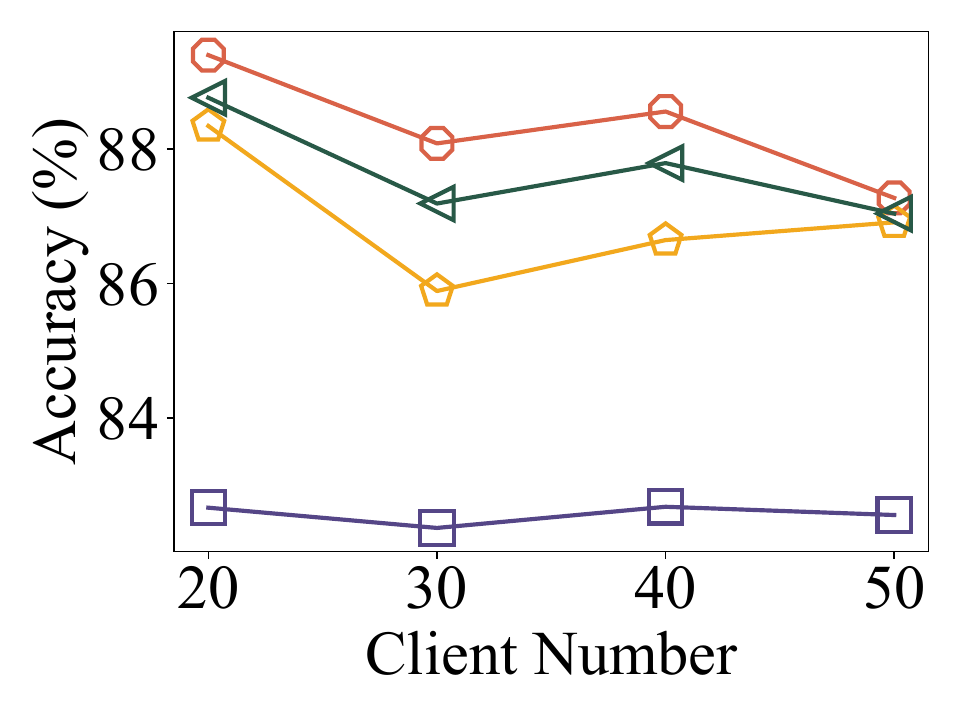}
        \caption{SNLI}
        \label{fig:snli_var_client_num}
    \end{subfigure}
    \begin{subfigure}[t]{0.47\linewidth}
        \centering
        \includegraphics[width=\linewidth]{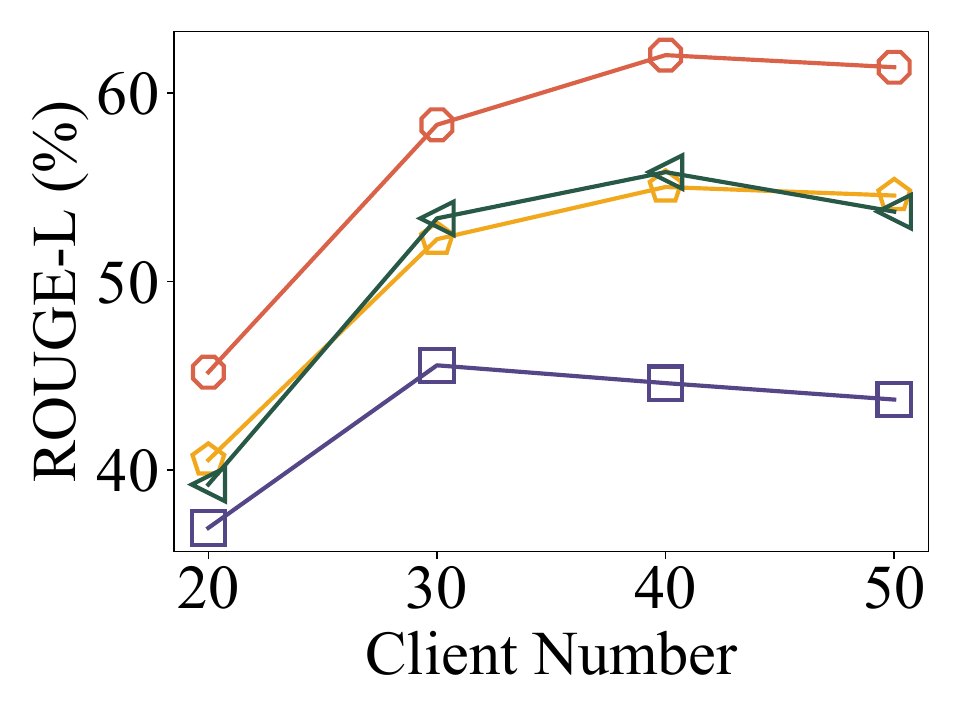}
        \caption{Natural Instructions}
        \label{fig:natural_instruct_var_client_num}
    \end{subfigure}
    \begin{subfigure}[t]{\linewidth}
        \vspace{0.5em}
        \centering
        \includegraphics[width=\linewidth]{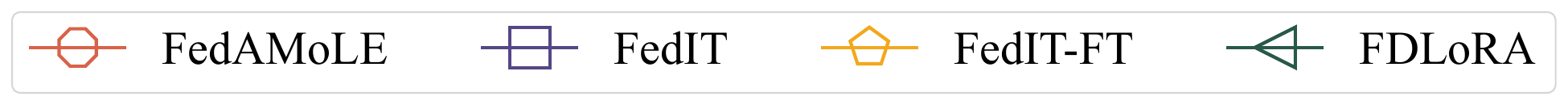}
    \end{subfigure}
    \vspace{-0.25cm}
    \caption{MTAL with different numbers of clients.}
    \vspace{-0.15cm}
    \label{fig:acc_var_client_num}
\end{figure}

Additionally, as shown in Figure~\ref{fig:acc_per_round}, FedAMoLE converges quickly across four typical non-IID scenarios and consistently outperforms other approaches during training. This indicates that even if training is stopped early for efficiency, FedAMoLE still achieves better results, demonstrating its practicality.
This advantage comes from FedAMoLE’s ability to dynamically adjust expert assignments, 
allowing it to respond promptly to changes in relevance between experts and client data as training progresses.
Furthermore, Figure~\ref{fig:acc_per_round} shows a clear performance hierarchy based on the degree of personalization: moving from vanilla FL to parameter-level personalization (via post-aggregation fine-tuning) yields significant gains, while FedAMoLE's architecture-level personalization achieves the best results across all four non-IID scenarios. This underscores that a greater degree of model personalization is key to superior performance on heterogeneous data.

Figure~\ref{fig:acc_var_client_num} shows the effect of client count. 
Notably, FedAMoLE consistently outperforms the strongest baselines under both label- and task-skewed settings, demonstrating its strong scalability.
\begin{figure*}[t]
    \centering
    \begin{subfigure}[t]{0.985\linewidth}
        \centering
        \includegraphics[width=\textwidth]{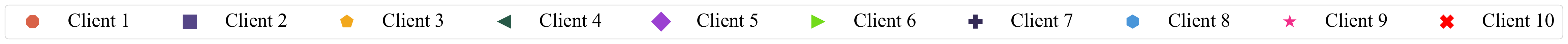}
    \end{subfigure}
    \begin{subfigure}[t]{0.47\textwidth}
        \centering
        \includegraphics[width=\linewidth]{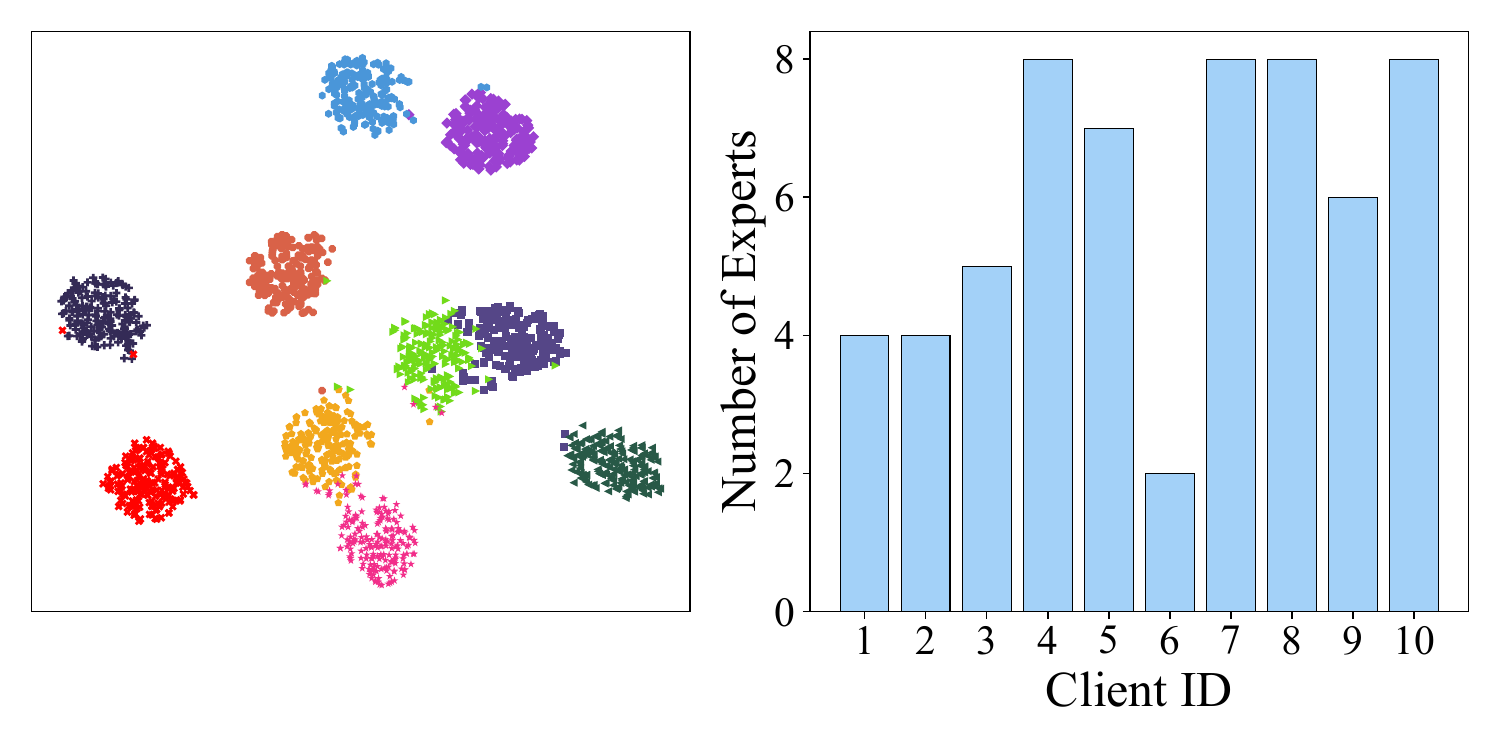}
        \caption{Results on SNLI ($\alpha=1.0$).}
        \label{fig:scatter_snli_round1_layer3_v_proj}
    \end{subfigure}
    \hspace{0.005\textwidth}
    \begin{subfigure}[t]{0.47\textwidth}
        \centering
        \includegraphics[width=\linewidth]{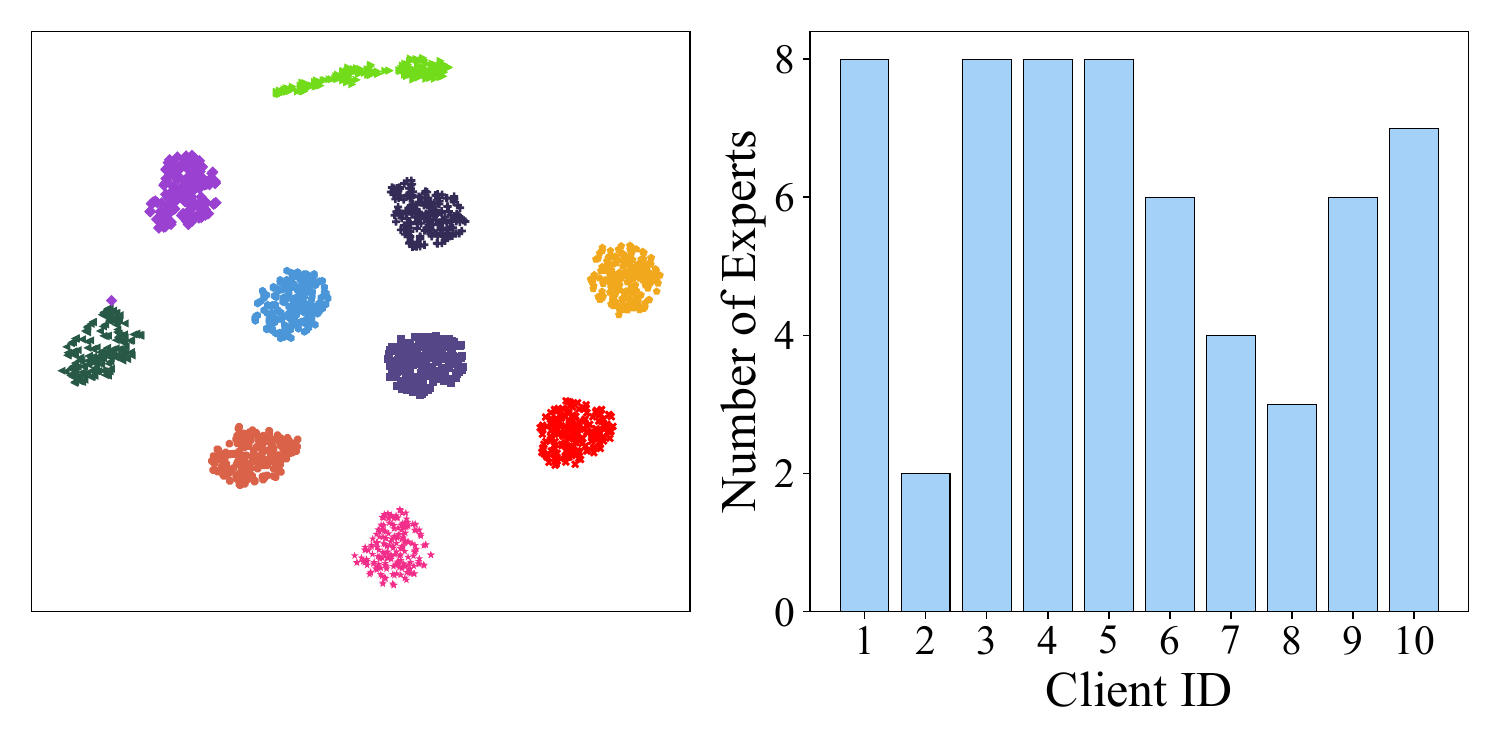}
        \caption{Results on Natural Instructions with one task per client.}
        \label{fig:scatter_natural_instruct_round1_layer3_v_proj}
    \end{subfigure}
    \caption{Visualization of the relevance between client data and domain experts in a HMoLE module after one round of FL. Each point in the scatter plot represents the t-SNE reduced relevance vector between a text sequence and all domain experts, with 200 points per client. The bar chart illustrates the number of domain experts assigned to each client in the next round.}
    \label{fig:logits_scatter}
\end{figure*}

\subsection{Comparison of Overhead} \label{sec:exp_system}
\begin{table}[t]
    \caption{Comparison of per-client system overhead in one FL round on the SNLI dataset using the LLaMA-3.2-1B backbone. ``Mem'' is the peak client-side GPU memory usage. ``Up'' and ``Down'' are the data sizes for client-to-server and server-to-client transfers, respectively. ``Train,'' ``Aggr,'' and ``  Test'' are the latencies for local training, server-side aggregation, and local testing. FedAMoLE (X) denotes a total of X experts per module; FedAMoLE$^*$ indicates an optimized parallel implementation; ``+'' denotes the overhead relative to FedIT-FT.}
    \small
    \centering
    \addtolength{\tabcolsep}{-4pt}
    \begin{tabular}{l|llllll}
        \toprule
        Approaches & Mem/GB & Up/MB & Down/MB & Train/s & Aggr/s & Test/s \\
        \midrule
        FedIT-FT & 2.78 & 1.64 & 1.64 & 4.85 & 0.02 & 2.37 \\
        FedPrompt-FT & 2.83 & 0.29 & 0.29 & 4.47 & 0.00 & 3.20 \\
        FedPTuning-FT & 2.84 & 0.16 & 0.16 & 4.19 & 0.00 & 2.99 \\
        CoMiGS & 8.00 & 21.55 & 21.55 & 52.67 & 0.04 & 19.21 \\
        FDLoRA & 2.78 & 1.64 & 1.64 & 4.14 & 0.03 & 2.60 \\
        \midrule
        FedAMoLE (15) &
        2.79\textsubscript{\textcolor{red!70!black}{\scriptsize +0.01}} & 
        7.59\textsubscript{\textcolor{red!70!black}{\scriptsize +6.0}} & 
        7.59\textsubscript{\textcolor{red!70!black}{\scriptsize +6.0}} & 
        13.25\textsubscript{\textcolor{red!70!black}{\scriptsize +8.4}} & 
        4.97\textsubscript{\textcolor{red!70!black}{\scriptsize +5.0}} &
        4.92\textsubscript{\textcolor{red!70!black}{\scriptsize +2.6}} \\
        FedAMoLE (20) &
        2.79\textsubscript{\textcolor{red!70!black}{\scriptsize +0.01}} & 
        9.24\textsubscript{\textcolor{red!70!black}{\scriptsize +7.6}} & 
        9.23\textsubscript{\textcolor{red!70!black}{\scriptsize +7.6}} & 
        14.44\textsubscript{\textcolor{red!70!black}{\scriptsize +9.6}} & 
        5.20\textsubscript{\textcolor{red!70!black}{\scriptsize +5.2}} & 
        5.20\textsubscript{\textcolor{red!70!black}{\scriptsize +2.8}} \\
        FedAMoLE (25) & 
        2.81\textsubscript{\textcolor{red!70!black}{\scriptsize +0.03}} & 
        10.88\textsubscript{\textcolor{red!70!black}{\scriptsize +9.2}} & 
        10.87\textsubscript{\textcolor{red!70!black}{\scriptsize +9.2}} & 
        15.57\textsubscript{\textcolor{red!70!black}{\scriptsize +10.7}} & 
        5.49\textsubscript{\textcolor{red!70!black}{\scriptsize +5.5}} & 
        5.63\textsubscript{\textcolor{red!70!black}{\scriptsize +3.3}} \\
        FedAMoLE (30) & 
        2.80\textsubscript{\textcolor{red!70!black}{\scriptsize +0.02}} & 
        12.52\textsubscript{\textcolor{red!70!black}{\scriptsize +10.9}} & 
        12.51\textsubscript{\textcolor{red!70!black}{\scriptsize +10.9}} & 
        16.49\textsubscript{\textcolor{red!70!black}{\scriptsize +11.6}} & 
        5.85\textsubscript{\textcolor{red!70!black}{\scriptsize +5.8}} & 
        5.72\textsubscript{\textcolor{red!70!black}{\scriptsize +3.4}} \\
        FedAMoLE$^*$ (30) & 
        3.55\textsubscript{\textcolor{red!70!black}{\scriptsize +0.77}} & 
        12.52\textsubscript{\textcolor{red!70!black}{\scriptsize +10.9}} & 
        12.51\textsubscript{\textcolor{red!70!black}{\scriptsize +10.9}} & 
        12.40\textsubscript{\textcolor{red!70!black}{\scriptsize +7.6}} & 
        5.10\textsubscript{\textcolor{red!70!black}{\scriptsize +5.1}} & 
        2.82\textsubscript{\textcolor{red!70!black}{\scriptsize +0.5}} \\
        \bottomrule
    \end{tabular}
    \label{tab:system_overhead}
\end{table}

Table~\ref{tab:system_overhead} presents the system efficiency of various PFL baselines and FedAMoLE. The RSEA process latency is included in FedAMoLE's aggregation time, while the black-box optimization for FDLoRA is part of its testing latency. 
We observe that despite employing multiple experts, FedAMoLE’s memory usage remains comparable to single LoRA-based baselines and is nearly unaffected by the total number of experts, since each expert is a lightweight LoRA adapter with negligible parameters compared to the pre-trained model. 
Even with 30 experts per module, the per-client data transfer is only 12.52 MB. 
With an optimized implementation, FedAMoLE achieves substantial reductions in both memory and computational overhead compared to another MoLE-based approach, CoMiGS.

FedAMoLE also maintains low per-round training (16.49 s) and inference (5.72 s) latencies even when fine-tuning a frozen 1B-parameter LLM with 30 experts.
The increased latency compared to non-MoE baselines is attributed to the additional overhead from RSEA strategy and the default memory-efficient implementation of the HMoLE module, which processes domain experts sequentially. 
However, this latency can be mitigated by parallelizing these computations at the cost of a minor increase in memory, as demonstrated by FedAMoLE$^*$. We expect such overheads to diminish further with future MoE hardware and software optimizations.

\begin{table}[t]
    \centering
    \begin{minipage}[t]{0.47\linewidth}
        \centering
        \captionof{table}{System overhead versus FedMoE. Setup and notation follow Table~\ref{tab:system_overhead}.}
\label{tab:switch_transformer_overhead}
\centering
\addtolength{\tabcolsep}{-4pt}
\begin{tabular}{c|ccc}
    \toprule
    Metric & FedMoE & FedAMoLE \\
    \midrule
    Mem/GB & 13.89 & 7.99 \\
    Up/MB & 1850 & 25.66 \\
    Down/MB & 1850 & 25.63 \\
    \bottomrule
\end{tabular}
    \end{minipage}
    \hfill
    \begin{minipage}[t]{0.49\linewidth}
        \centering
        \captionof{table}{RSEA expert assignment time (s) on a single Intel Core i5-13400F CPU.}
\label{tab:opt_overhead}
\centering
\addtolength{\tabcolsep}{-2pt}
\begin{tabular}{c|ccc}
    \toprule
    \diagbox[width=6em,height=2.5ex,innerleftsep=0.5pt,innerrightsep=0.5pt]{\raisebox{-0.8ex}[0pt][0pt]{\small Client}}{\raisebox{0.8ex}[0pt][0pt]{\small Expert}} & 10 & 50 & 100 \\
    \midrule
    30 & 0.57 & 2.17 & 4.37 \\
    50 & 0.90 & 3.45 & 6.87 \\
    100 & 1.63 & 7.62 & 15.28 \\
    \bottomrule
\end{tabular}
    \end{minipage}
\end{table}

To further demonstrate efficiency, we compare FedAMoLE with MoE-based methods that utilize dense experts. Since FedMoE~\cite{meiFedMoEPersonalizedFederated2024} is not open-source, we cannot directly measure its performance on our LLaMA-based setup. To ensure a fair comparison, we aligned FedAMoLE with its original experimental setting (a Switch Transformer backbone on the AG News dataset) and compared our results with the data reported in their paper. As shown in Table~\ref{tab:switch_transformer_overhead}, FedAMoLE leverages LoRA's parameter efficiency to compress the per-round communication cost to just 25.66 MB (approximately 1\% of FedMoE's reported GB-level volume) and reduces memory usage by nearly half, making federated LLM fine-tuning practical in real-world scenarios.

Finally, we evaluate the scalability of RSEA strategy. 
From Table~\ref{tab:opt_overhead}, the primary overhead of RSEA—expert assignment—takes only 15s even with 100 clients and 100 experts—a typical upper bound for cross-silo scenarios~\cite{chaoyanghe2020fedml}. Considering that the per-round time is dominated by local training and communication, our approach is highly practical for typical cross-silo FL deployments while the underlying MILP problem is $\mathcal{N}\mathcal{P}$-hard in the worst case.

\subsection{Ablation Study} \label{sec:ablation}
To assess the contributions of the proposed HMoLE module, RSEA strategy, and shared expert, we design four ablation variants: (1) FedAMoLE-H, which replaces HMoLE with vanilla MoLE (\ref{eq:mole_routing})—each router uses a fixed output dimension equal to the total number of assignable experts to ensure aggregation compatibility, with top-$k^e$ routing restricted to assigned experts; (2) FedAMoLE-S, where the shared expert is ablated, and only domain expert collaboration is used; (3) FedAMoLE-R, where the RSEA strategy is removed and experts are manually assigned in the first round, remaining fixed thereafter; (4) FedAMoLE (Random), where the RSEA strategy is removed and experts are randomly assigned in each round.

\begin{table}[t]
    \caption{Ablation study conducted on SNLI and NI.
    }
    \vspace{-0.2cm}
    \label{tab:ablation}
    \centering
    \begin{tabular}{c|c|ccccc}
        \toprule
        \multirow{2.5}{*}{Dataset} & FedIT & \multicolumn{5}{c}{FedAMoLE} \\
        \cmidrule(lr){2-7}
        & +FT & -H & -S & -R & Random & \\
        \midrule
        SNLI & 86.45 & 86.90 & 87.50 & 87.20 & 88.05 & \textbf{88.95} \\
        NI & 53.45 & 59.57 & 56.04 & 60.82 & 52.92 & \textbf{61.29} \\
        \bottomrule
    \end{tabular}
    \vspace{-0.6cm}
\end{table}
As presented in Table~\ref{tab:ablation}, FedAMoLE-R significantly outperforms FedIT-FT in accuracy, demonstrating the benefits of heterogeneous models. FedAMoLE achieves superior accuracy over FedAMoLE-R, highlighting the advantages of data-driven model architectures. The performance degradation of both FedAMoLE-H and FedAMoLE (Random), relative to FedAMoLE, underscores the importance of the HMoLE module and the RSEA strategy. Moreover, the inferior performance of FedAMoLE-S compared to FedAMoLE emphasizes the necessity of shared experts for general knowledge sharing, rather than relying solely on domain experts.

\subsection{Discussion on Expert Assignment} \label{sec:discuss_expert_assign}
Figure~\ref{fig:logits_scatter} shows that RSEA adaptively allocates domain experts based on data distribution. In the scatter plot, each point represents the t-SNE-reduced relevance vector between a text sequence and all domain experts. This vector is calculated via (\ref{eq:client_embedding}-\ref{eq:hmole_relevance}), where $\mathcal{D}^{\text{emb}}_i$ in (\ref{eq:client_embedding}) contains only one sample. Points clustered near the origin signify uniform relevance, indicating an alignment with the global distribution that requires fewer domain experts. Conversely, points distant from the origin show a strong affinity for specific experts, indicating more complex, specialized data that necessitates a greater model capacity provided by more domain experts. This pattern is clearly reflected in FedAMoLE’s expert assignments. For example, in Figure~\ref{fig:scatter_natural_instruct_round1_layer3_v_proj}, the data points for clients 2, 7, and 8 are concentrated near the origin, indicating their local data aligns with the global distribution. As a result, the corresponding bar chart confirms that these clients are assigned fewer experts. Please refer to \S\ref{sec:rsea_stability} for further discussion on the stability of expert assignment.
\section{Conclusion}
This work introduces FedAMoLE, a novel PFL framework for LLM fine-tuning, characterized by a heterogeneous MoLE module to enable architectural model heterogeneity, and a reverse selection-based expert assignment strategy to optimize model architectures in a data-driven manner.
It enables architecture-level model personalization across clients with low communication overhead.
Extensive experiments on seven scenarios demonstrate that FedAMoLE achieves superior accuracy over existing methods in scenarios with non-IID data, highlighting the potential of data-driven optimization for model architectures in LLM fine-tuning with FL.


\bibliographystyle{ACM-Reference-Format}
\bibliography{reference}

\appendix

\section{More Related Work}
\label{appendix:related-work}
\subsection{Federated Fine-Tuning for LLMs}
Current federated fine-tuning approaches for LLMs primarily focus on improving system efficiency to handle the significant resource demands of these models. To reduce communication overhead, many methods adopt PEFT techniques~\cite{zhangBuildingFederatedgptFederated2023, cheFederatedLearningLarge2023} or employ Zero-Order Optimization (ZOO) to drastically compress model updates~\cite{qinFederatedFullParameterTuning2023b}. To lower on-device computational and memory footprints, other approaches investigate methods such as quantization~\cite{xu2024fwdllm} or the fine-tuning of compressed LLMs~\cite{wuFedBiOTLLMLocal2024}. While these studies have significantly advanced the ecosystem and minimized overheads, they overlook the challenge of data heterogeneity, which causes performance degradation in practical FL applications~\cite{tanPersonalizedFederatedLearning2023}.

\subsection{Personalized Federated Learning} \label{sec:pfl_related_work}
Personalized Federated Learning (PFL) has been widely studied to address data heterogeneity. Many approaches focus on parameter-level personalization by adapting model parameters to client-specific distributions through techniques such as fine-tuning~\cite{yuSalvagingFederatedLearning2022}, regularization~\cite{liFederatedOptimizationHeterogeneous},  clustering~\cite{ghoshEfficientFrameworkClustered2021}, and parameter decoupling~\cite{collinsExploitingSharedRepresentations2021}. Although these methods help mitigate performance degradation, their homogeneous model architectures limit personalization capability. For greater adaptability, some works explore architecture-level personalization using knowledge distillation~\cite{linEnsembleDistillationRobust2020} and ensemble learning~\cite{qin2023fedapen}. However, knowledge distillation often requires additional data (e.g., public datasets), which contradicts the federated learning goal of addressing data scarcity. Meanwhile, ensemble learning necessitates training multiple models locally—an impractical solution given the high resource cost of training a single LLM.

\subsection{Personalized Federated Learning with Mixture of Experts}
Mixture-of-Experts (MoE) architectures have attracted attention in PFL due to their ability to achieve model heterogeneity while maintaining constant computational overhead. Early MoE-based PFL research focused on small-scale models like CNNs and RNNs, often treating the entire model as a single expert. Some works~\cite{petersonPrivateFederatedLearning2019,yiPFedMoEDataLevelPersonalization2024} combined a shared expert with a personalized expert using a gating network. Others supported mixing multiple experts through personalized weighting coefficients~\cite{reisserFederatedMixtureExperts2021} or client selection~\cite{dunFedJETsEfficientJusttime2023}. However, these works are primarily designed for small-scale neural networks and incur significant memory and communication overhead, making them unsuitable for LLMs. While a recent effort adopts lightweight experts to improve training efficiency~\cite{almansooriCollaborativeEfficientPersonalization2024}, it still restricts personalization to the parameter level.

\section{Preliminaries} \label{sec:preliminaries}
\subsection{Mixture of Experts and A Lightweight Improvement}
MoE~\cite{jacobsAdaptiveMixturesLocal1991, fedusSwitchTransformersScaling2021} enables the scaling of LLMs with constant computational overhead. 
A typical MoE layer consists of a linear router and multiple sub-modules termed ``experts''. Given a specific input, the router sparsely selects a fixed number of experts based on its features. This process can be formalized as
\begin{equation}
    \begin{aligned}
        &p_i=\text{softmax}(\mathbf{W}_r \mathbf{x})_i\\
        &\mathcal{T}=\text{top-}k(p_i)\\
        &\mathbf{y}=\sum_{i\in\mathcal{T}}p_i E_i(\mathbf{x}),
        \label{eq:moe_routing}
    \end{aligned}
\end{equation}
where $\mathbf{x}\in\mathbb{R}^d$ represents the input features, $\mathbf{W}_r\in\mathbb{R}^{N\times d}$ is the parameter weights of the router (with $N$ experts), $p_i$ represents the weighting coefficient of expert $i$ for input $\mathbf{x}$, $\mathcal{T}$ is the set of indices of the $k$ selected experts, $E_i$ is the $i$-th expert sub-module (such as a two-layer feed-forward neural network), and $\mathbf{y}$ is the final output.

The adaptive collaboration of domain experts in MoE enables effective adaptation to diverse data distributions, making it a promising solution for personalizing client models in FL. However, applying MoE to federated LLM fine-tuning incurs significant communication costs due to the transfer of dense expert sub-modules between the server and clients. A natural solution to this limitation is to make MoE experts more lightweight.

Mixture-of-LoRA-Experts (MoLE) architectures~\cite{liMixLoRAEnhancingLarge2024, gaoHigherLayersNeed2024} address this by using LoRA adapters as experts within the MoE framework.

Low-rank adaptation (LoRA)~\cite{hu2022lora} is a widely adopted Parameter-Efficient Fine-Tuning (PEFT) technique for LLMs. 
It assumes that the updates $\Delta \mathbf{W}$ to the pre-trained parameters $\mathbf{W} \in \mathbb{R}^{h \times d}$ are highly sparse and can be approximated by the product of two low-rank matrices $\mathbf{A} \in \mathbb{R}^{r \times d}$ and $\mathbf{B} \in \mathbb{R}^{h \times r}$, i.e., LoRA adapters, with $r \ll \min(d, h)$ to minimize parameter size. The fine-tuned parameters $\mathbf{W}$ during forward computation can then be expressed as:
\begin{equation}
    \mathbf{y}=\mathbf{Wx}+\Delta \mathbf{Wx}=\mathbf{Wx+BAx}.
\end{equation}

MoLE substantially reduces the scale of MoE experts with the lightweight nature of LoRA adapters.
The forward computation of MoLE is given by:
\begin{equation}
    \begin{aligned}
        &p_i=\text{softmax}(\mathbf{W}_r \mathbf{x})_i\\
        &\mathcal{T}=\text{top-}k(p_i)\\
        &\mathbf{y}=\mathbf{Wx}+\sum_{i\in\mathcal{T}}p_i \mathbf{B}_i\mathbf{A}_i\mathbf{x}.
    \end{aligned}
    \label{eq:mole_routing}
\end{equation}

\subsection{Limitations of Vanilla MoLE for FL}
Assigning varying numbers of domain experts to instances of a traditional MoLE module across clients challenges federated aggregation. Consider the following scenario: if two client instances of a MoLE module are assigned to three and five domain experts, respectively, the corresponding router dimensions would be $3 \times d$ and $5 \times d$, as defined in (\ref{eq:mole_routing}). This dimensional inconsistency prevents direct aggregation of routers via FedAvg~\cite{mcmahan2017fl}. The core issue originates that the parameter dimension of a traditional MoLE router is tied to the number of domain experts, resulting in shape mismatches across module instances with varying domain expert counts.

\section{More Experiments}
\subsection{Implementation Details} \label{sec:impl}
Methods are implemented with \texttt{PyTorch 2.4.0}~\cite{paszkePyTorchImperativeStyle2019}, \texttt{PEFT}~\cite{peft} and \texttt{Transformers}~\cite{wolfHuggingFacesTransformersStateart2020}, and performed on a server with NVIDIA RTX 4090 GPU using \texttt{BF16} precision, unless stated otherwise.

Our Hybrid dataset is a composite of ten datasets for different tasks, including Dolly-15K~\cite{DatabricksBlog2023DollyV2}, Alpaca~\cite{alpaca}, NI~\cite{wangSuperNaturalInstructionsGeneralizationDeclarative2022}, SNLI~\cite{bowmanLargeAnnotatedCorpus2015}, AG News, Yelp-P~\cite{zhang2015character}, and four sub-tasks from SuperGLUE~\cite{wangSuperGLUE2019}. For all scenarios, each client's local data is randomly split into training (80\%), validation (10\%), and test (10\%) sets. Due to computational constraints, we cap all validation sets at 200 samples, and test sets at 200 for SNLI and 50 for the remaining datasets.

Unless otherwise specified, all methods use Adam optimizer with a batch size of 1, a learning rate $\eta = 5e^{-5}$, and a learning rate decay of 0.99 per round. For LoRA-based approaches, we set $r=8$, $\alpha=16$, and a dropout rate of 5\%, applying LoRA to $\mathbf{Q}$ and $\mathbf{V}$ in self-attention blocks. For Prompt Tuning, the virtual prompt is initialized using the soft prompt for Dolly-15K, NI, and SNLI, while for Hybrid and BigBench, it is randomly initialized with 20 virtual tokens. For P-Tuning, we set the virtual token count to 20 and use an MLP as the prompt encoder. In FDLoRA, the InnerOpt phase synchronizes the shared and personalized adapters every 5 rounds; the OuterOpt phase uses a learning rate of 1.0 and a momentum of 0.5; the FusionOpt phase uses \texttt{nevergrad}~\cite{nevergrad} as the black-box optimizer with an L1 regularization weighted by 0.05. 
For CoMiGS, the total number of experts is kept the same as in this work for a fair comparison, while other parameters follow the recommendations in the corresponding paper\cite{fanDeviceCollaborativeLanguage2024}.
For FedAMoLE, we use SCIP~\cite{BolusaniEtal2024OO} to solve the optimization problem, setting the total assignable experts per module to 30, with $k^e = 2$, $k^c = 2$, $b = 8$, and $\beta = 1e^{-3}$.

\subsection{Comparison of Accuracy on Additional LLM Sizes and Backbones} \label{sec:compare_acc_extended}
\begin{table*}[t]
    \caption{
    Comparison of accuracy on LLaMA-3.2-3B and Qwen-2.5-0.5B. Notation follows Table~\ref{tab:compare_acc}.
    }
    \vspace{-0.25cm}
    \setlength\tabcolsep{3.8pt}
    \centering
    \begin{tabular}{c|l|c|c|c|ccc|c|c}
        \toprule
        \multicolumn{2}{c|}{\multirow{2}{*}{Approach}} & \textbf{BigBench} & \textbf{Hybrid} & \textbf{NI} & \multicolumn{3}{c|}{\textbf{Dolly-15K}} & \textbf{SNLI} & \multirow{2}{*}{Average} \\
        \multicolumn{2}{c|}{} & 1 domain/client & 1 task/client & 1 task/client & $\alpha=0.1$ & $\alpha=1.0$ & $\alpha=100.0$ & $\alpha=1.0$ & \\
        \midrule
        \multirow{5.5}{*}{\rotatebox{90}{LLaMA-3.2-3B}} & FedIT & 54.28$\pm$0.72 & 64.14$\pm$0.62 & 64.80$\pm$0.94 & \underline{33.10$\pm$0.54} & 31.17$\pm$1.15 & \underline{29.72$\pm$0.49} & 88.02$\pm$0.50 & 52.18 \\
        & FedIT-FT & \underline{56.22$\pm$1.32} & 65.17$\pm$0.40 & \underline{64.97$\pm$1.47} & 32.35$\pm$1.38 & \underline{31.51$\pm$1.11} & 29.40$\pm$0.19 & 90.18$\pm$1.26 & \underline{52.83} \\
        & FDLoRA & 55.42$\pm$2.37 & \underline{66.38$\pm$0.54} & 63.30$\pm$1.20 & 30.82$\pm$1.13 & 29.45$\pm$1.59 & 28.49$\pm$1.17 & \underline{90.50$\pm$1.06} & 52.05 \\
        \cmidrule(lr){2-10}
        & FedAMoLE & \textbf{57.71$\pm$0.35} & \textbf{68.55$\pm$1.70} & \textbf{66.65$\pm$1.06} & \textbf{33.44$\pm$1.35} & \textbf{31.61$\pm$1.40} & \textbf{30.55$\pm$0.38} & \textbf{91.17$\pm$0.56} & \textbf{54.24} \\
        & Gains & 2.65 & 3.26 & 2.58 & 1.02 & 0.33 & 2.77 & 0.73 & 2.67 \\
        \midrule
        \multirow{5.5}{*}{\rotatebox{90}{Qwen-2.5-0.5B}} & FedIT & 39.69$\pm$0.23 & 60.38$\pm$0.32 & 50.32$\pm$1.82 & \underline{28.82$\pm$0.59} & \underline{28.42$\pm$1.72} & 26.98$\pm$0.59 & 83.28$\pm$1.05 & 45.41 \\
        & FedIT-FT & 42.55$\pm$0.67 & 59.88$\pm$0.46 & \underline{53.50$\pm$1.31} & \underline{28.82$\pm$0.67} & 28.04$\pm$1.61 & 26.59$\pm$0.55 & \textbf{87.80$\pm$0.55} & \underline{46.74} \\
        & FDLoRA & \underline{43.07$\pm$1.64} & \underline{60.82$\pm$0.79} & 51.02$\pm$1.40 & 28.32$\pm$1.43 & 27.94$\pm$1.22 & \underline{27.37$\pm$0.91} & 86.72$\pm$1.36 & 46.46 \\
        \cmidrule(lr){2-10}
        & FedAMoLE & \textbf{49.34$\pm$1.93} & \textbf{61.65$\pm$0.87} & \textbf{60.31$\pm$1.11} & \textbf{29.08$\pm$0.97} & \textbf{29.04$\pm$1.00} & \textbf{27.54$\pm$0.17} & \underline{87.66$\pm$0.57} & \textbf{49.23} \\
        & Gains (\%) & 14.55 & 1.38 & 12.74 & 0.90 & 2.18 & 0.60 & -0.16 & 5.33 \\
        \bottomrule
    \end{tabular}
    \vspace{-0.35cm}
    \label{tab:compare_acc_extended}
\end{table*}
To further validate the effectiveness of FedAMoLE, we compare FedAMoLE to strong baselines with additional LLM sizes (LLaMA-3.2-3B\footnote{Adopt checkpoint from \url{https://huggingface.co/meta-llama/Llama-3.2-3B}.}) and backbones (Qwen-2.5-0.5B\footnote{Adopt checkpoint from \url{https://huggingface.co/Qwen/Qwen2.5-0.5B}.}).
Results in Table~\ref{tab:compare_acc_extended} show the consistently superior performance of FedAMoLE.

\subsection{Hyper-parameter Sensitivity} \label{sec:hyper_sens}
\begin{figure}[H]
    \centering
    \begin{subfigure}[t]{0.5\linewidth}
        \centering
        \includegraphics[width=\linewidth]{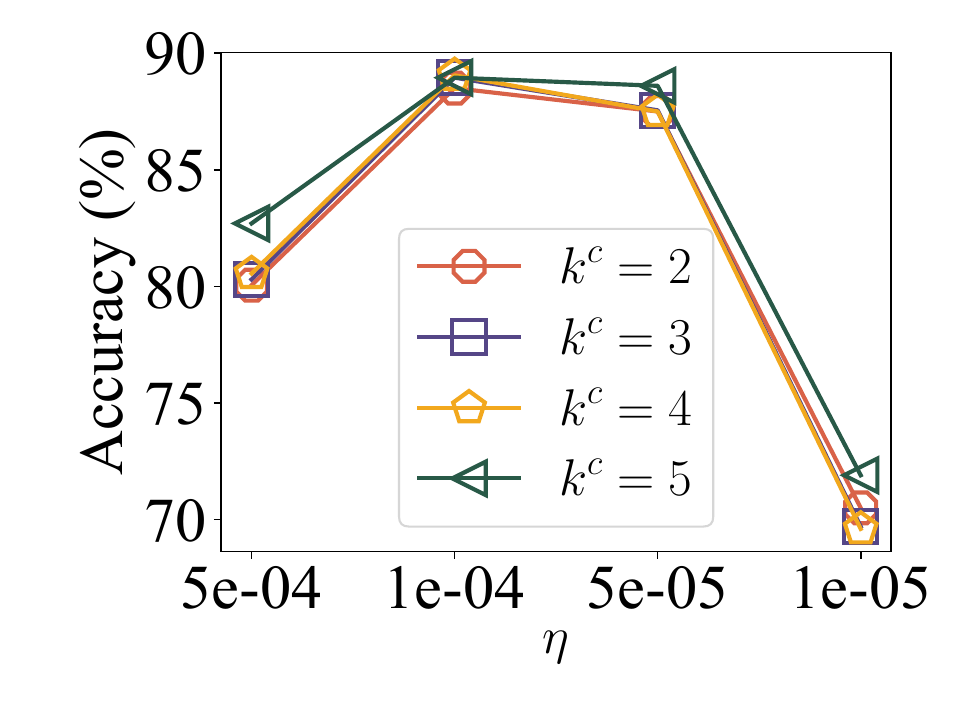}
        \caption{The impact on accuracy by the learning rate $\eta$ and the number of clients selected by each expert $k^c$.}
        \label{fig:expert_choices_vs_lr}
    \end{subfigure}
    \hspace{0.01\linewidth}
    \begin{subfigure}[t]{0.47\linewidth}
        \centering
        \includegraphics[width=\linewidth]{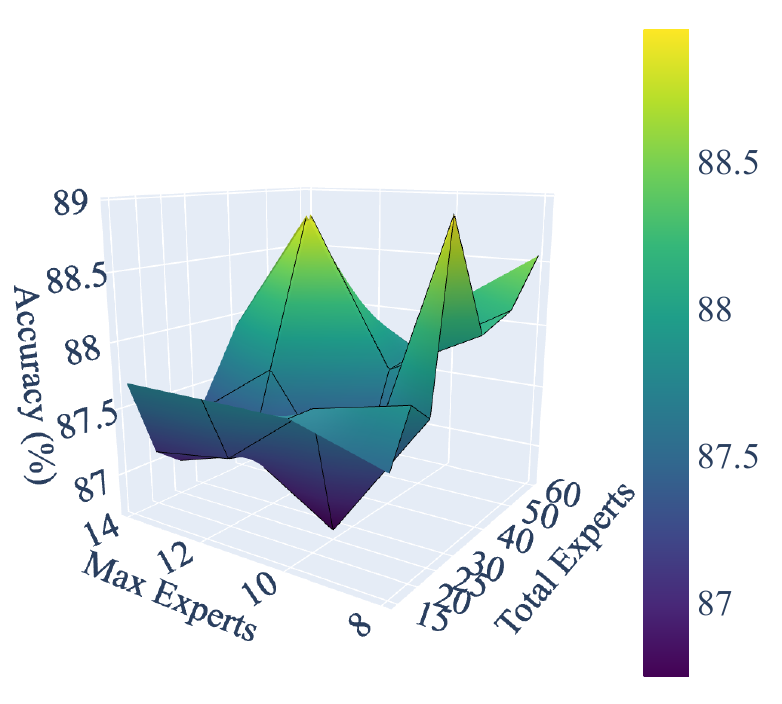}
        \caption{The impact by the number of total experts and maximum experts per module $b$ on accuracy.}
        \label{fig:num_experts_vs_max_experts}
    \end{subfigure}
    \caption{Hyperparameter sensitivity of FedAMoLE.}
    \label{fig:hyper_params_sens}
\end{figure}
Figure \ref{fig:expert_choices_vs_lr} presents the hyperparameter sensitivity of FedAMoLE evaluated on SNLI, showing that the performance of FedAMoLE is minimally affected by the number of clients each expert selects. However, over-large or over-small the learning rate declines the performance of FedAMoLE, indicating the need for an appropriate learning rate to ensure optimal performance.

Figure~\ref{fig:num_experts_vs_max_experts} illustrates the impact of the total number of experts per module and the maximum number of experts per client on FedAMoLE's performance. Overall, these two parameters have a limited impact, with model accuracy fluctuating within a narrow 2\% range, demonstrating FedAMoLE's robustness to hyperparameter settings. As the figure shows, the model's performance peaks with 30 total experts and $b=8$. Therefore, we adopt this configuration as the default set of hyperparameters for our experiments.

\subsection{Stability of Expert Assignment} \label{sec:rsea_stability}
\begin{table}[H]
    \vspace{-0.4cm}
    \caption{Expert assignment of an HMoLE module on a client in the last 5 rounds. Columns indicate training rounds, rows indicate expert indices, and “\checkmark” denotes assignment.}
    \vspace{-0.35cm}
    \label{tab:expert_assign_across_rounds}
    \scriptsize
    \centering
    \setlength{\tabcolsep}{1pt}
    \begin{tabular}{|c|c|c|c|c|c|c|c|c|c|c|c|c|c|c|c|c|c|c|c|c|c|c|c|c|c|c|c|c|c|c|}
        \cline{1-31}
        & 1 & 2 & 3 & 4 & 5 & 6 & 7 & 8 & 9 & 10 & 11 & 12 & 13 & 14 & 15 & 16 & 17 & 18 & 19 & 20 & 21 & 22 & 23 & 24 & 25 & 26 & 27 & 28 & 29 & 30 \\
        \cline{1-31}
        26 & & & & & & & & & \checkmark & \checkmark & \checkmark & & \checkmark & & & \checkmark & & \checkmark & & & & & & & & & & \checkmark & & \checkmark \\
        \cline{1-31}
        27 & & & & & & & \checkmark & & \checkmark & \checkmark & \checkmark & & \checkmark & & & \checkmark & & \checkmark & & & & & & & & & & & & \checkmark \\
        \cline{1-31}
        28 & & & & & & & & & & \checkmark & \checkmark & & \checkmark & & & \checkmark & & \checkmark & \checkmark & & & & & & & & & \checkmark & & \checkmark \\
        \cline{1-31}
        29 & & & & & & & & & & \checkmark & \checkmark & & \checkmark & & & \checkmark & & \checkmark & \checkmark & & & & & & & & & \checkmark & & \checkmark \\
        \cline{1-31}
        30 & & \checkmark & & & & & & & & \checkmark & \checkmark & & \checkmark & & & \checkmark & & \checkmark & & & & & & & & & & \checkmark & & \checkmark \\
        \cline{1-31}
    \end{tabular}
    \vspace{-0.45cm}
\end{table}
\begin{wraptable}[9]{r}{0.38\linewidth}
    \vspace{-0.48cm}
    \caption{Expert count for the HMoLE module in Table~\ref{tab:expert_assign_across_rounds} across different  random seeds.}
    \vspace{-0.35cm}
    \label{tab:expert_count_across_runs}
    \centering
    \addtolength{\tabcolsep}{-2pt}
    \renewcommand{\arraystretch}{0.9}
    \begin{tabular}{c|ccccc}
        \toprule
        seed & 26 & 27 & 28 & 29 & 30 \\
        \midrule
        42 & 8 & 8 & 8 & 8 & 8 \\
        62 & 8 & 8 & 8 & 7 & 8 \\
        82 & 7 & 7 & 8 & 6 & 8 \\
        \bottomrule
    \end{tabular}
\end{wraptable}
Empirical results show that expert assignment under the RSEA strategy remains stable across both runs and training rounds. We illustrate this using experiment logs from the NI dataset (Table~\ref{tab:compare_acc}). Specifically, we randomly select an HMoLE module on a client and track its expert assignments and expert counts over the final five rounds. As shown in Table~\ref{tab:expert_assign_across_rounds}, the assignments converge clearly, with experts 10, 11, 13, 16, 18, and 30 consistently selected and only minor fluctuations observed, demonstrating cross-round stability. Table~\ref{tab:expert_count_across_runs} further shows that the number of assigned experts per round is consistent across seeds, confirming cross-run stability. Note that we omit expert IDs here, as experts are initialized differently under each seed, making ID-level comparisons infeasible.

\subsection{Necessity of Personalization}
\begin{figure}[t]
    \centering
    \begin{subfigure}[t]{0.47\linewidth}
        \centering
        \includegraphics[width=\linewidth]{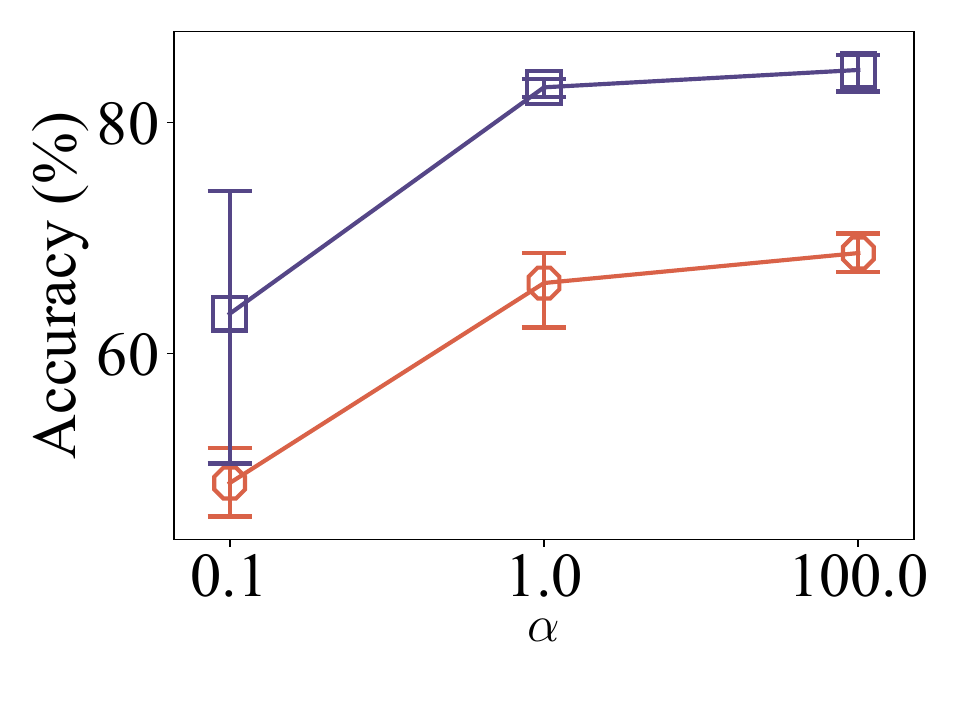}
        \vspace{-0.7cm}
        \caption{SNLI}
        \label{fig:snli_var_iids}
    \end{subfigure}
    \hspace{0.01\linewidth}
    \begin{subfigure}[t]{0.47\linewidth}
        \centering
        \includegraphics[width=\linewidth]{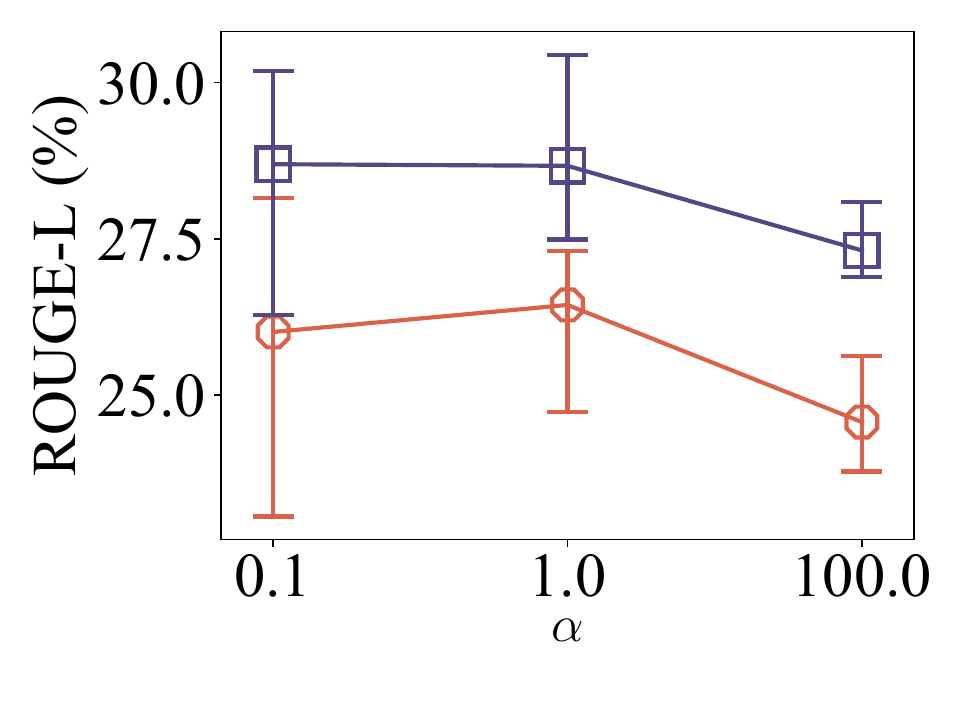}
        \vspace{-0.7cm}
        \caption{Dolly-15K}
        \label{fig:dolly_15k_var_iids}
    \end{subfigure}

    \vspace{0.1em}
    \begin{subfigure}[t]{0.5\linewidth}
        \centering
        \includegraphics[width=\linewidth]{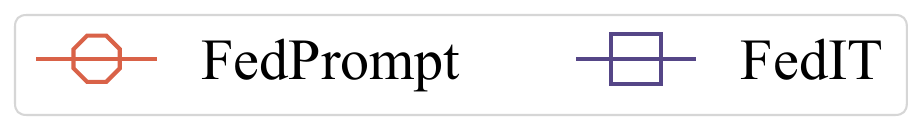}
    \end{subfigure}
    \vspace{-0.35cm}
    \caption{MTAL of vanilla FL approaches under varying data heterogeneity. Each data point represents the average result over three random seeds, with error bars showing the range between the best and worst values across the three seeds.}
    \label{fig:vanilla_fl_var_iids}
    \vspace{-0.1cm}
\end{figure}
Figure~\ref{fig:vanilla_fl_var_iids} illustrates the performance of two vanilla FL approaches across varying data heterogeneity. On SNLI, performance declines significantly as heterogeneity increases, aligning with existing FL works for relatively small models~\cite{tanPersonalizedFederatedLearning2023}. Our results confirm that LLM federated fine-tuning is similarly affected, underscoring the need for personalization mechanisms. In contrast, on Dolly-15K, accuracy remains stable despite data heterogeneity, likely because the dataset has fewer task types (8) with smaller inter-task differences, limiting the impact of heterogeneous partitioning.

\subsection{Robustness to Differential Privacy} \label{sec:dp_exp}
\begin{figure}[t]
    \centering
    \begin{subfigure}[t]{0.47\linewidth}
        \centering
        \includegraphics[width=\linewidth]{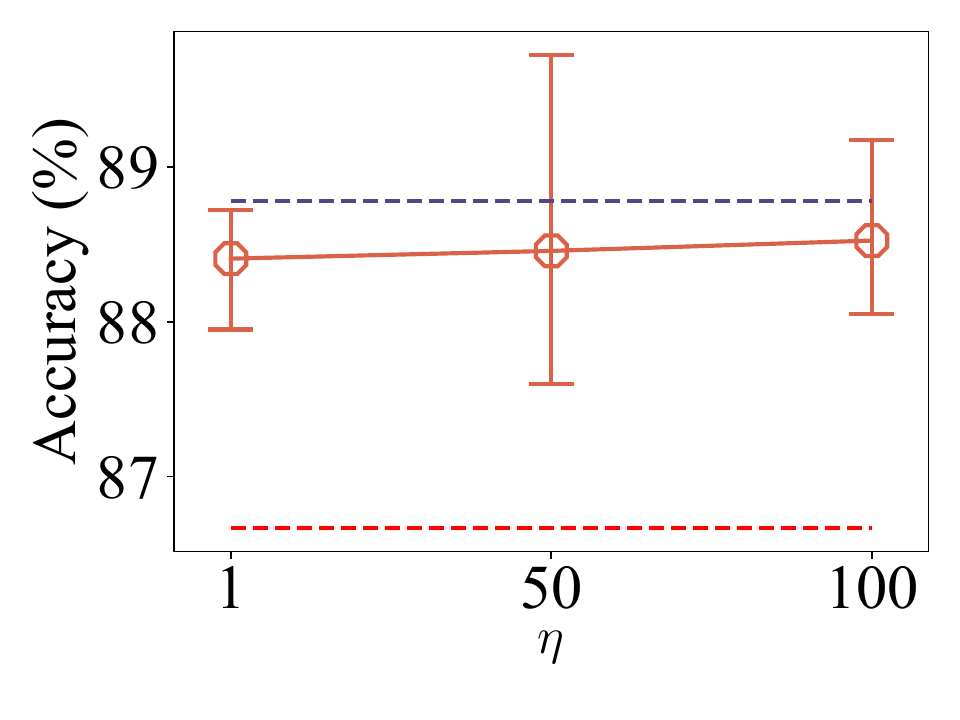}
        \vspace{-0.7cm}
        \caption{SNLI}
        \label{fig:snli_dp}
    \end{subfigure}
    \hspace{0.01\linewidth}
    \begin{subfigure}[t]{0.47\linewidth}
        \centering
        \includegraphics[width=\linewidth]{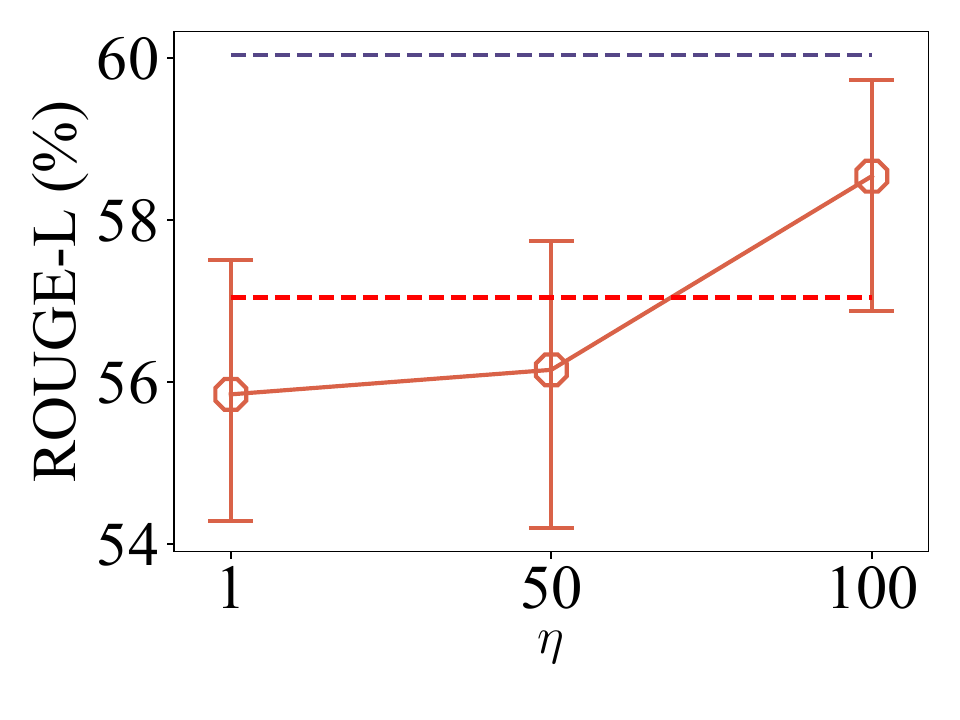}
        \vspace{-0.7cm}
        \caption{Natural Instructions}
        \label{fig:ni_dp}
    \end{subfigure}

    \vspace{0.1em}
    \begin{subfigure}[t]{0.94\linewidth}
        \centering
        \includegraphics[width=\linewidth]{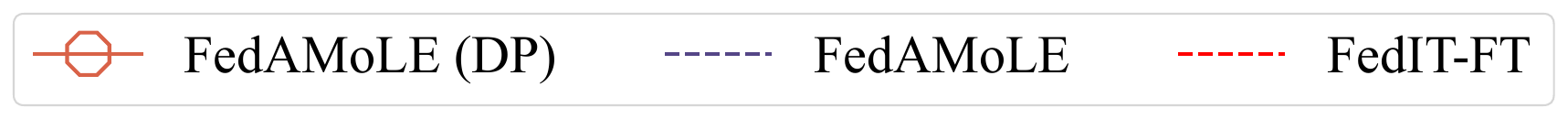}
    \end{subfigure}
    \vspace{-0.35cm}
    \caption{MTAL of FedAMoLE under varying privacy budgets $\eta$. Each point shows the average result over three seeds, with error bars indicating the min–max range. Dashed lines denote non-private baselines (averaged over three seeds).}
    \label{fig:dp_var_eta}
\end{figure}
Figure~\ref{fig:dp_var_eta} shows the accuracy of FedAMoLE under varying privacy budgets $\eta$ (\S\ref{sec:dp_method}). As expected, applying DP leads to performance degradation, as the added noise in expert and data embeddings hinders RSEA from accurately measuring expert–client relevance, resulting in suboptimal model architectures. This effect is more pronounced on NI, where greater data heterogeneity makes clients more sensitive to architectural mismatches. Despite the noisy embeddings, FedAMoLE still outperforms the strongest baseline FedIT-FT in most settings, except under high privacy budgets ($\eta=1, 50$) on NI, confirming its robustness to DP.

\end{document}